%% file: paper.tex
\documentclass{article}

\usepackage{microtype}
\usepackage{graphicx}
\usepackage{subfigure}
\usepackage{booktabs} 

\usepackage{hyperref}

\usepackage[accepted]{icml2024}

\usepackage{amsmath}
\usepackage{amssymb}
\usepackage{mathtools}
\usepackage{amsthm}

\usepackage[capitalize,noabbrev]{cleveref}

\theoremstyle{plain}
\newtheorem{theorem}{Theorem}[section]

\newtheorem{lemma}[theorem]{Lemma}
\newtheorem{corollary}[theorem]{Corollary}
\theoremstyle{definition}
\newtheorem{definition}[theorem]{Definition}
\newtheorem{assumption}[theorem]{Assumption}
\theoremstyle{remark}

\usepackage[textsize=tiny]{todonotes}
\input{math_commands.tex}

\usepackage{graphicx}
\usepackage{hyperref}
\usepackage{url}
\usepackage{verbatim}
\usepackage{amsthm}
\usepackage{wrapfig}
\usepackage{multirow}
\usepackage{booktabs}
\usepackage{array}
\usepackage{wrapfig}

\allowdisplaybreaks[2]
\def\ie{\textit{i.e.,~}}

\DeclareMathOperator*{\var}{Var}
\icmltitlerunning{Perfect Alignment May be Poisonous to Graph Contrastive Learning}

\begin{document}

\twocolumn[
\icmltitle{Perfect Alignment May be Poisonous to Graph Contrastive Learning}



\icmlsetsymbol{equal}{*}
\icmlsetsymbol{corresponding}{!}
\begin{icmlauthorlist}
\icmlauthor{Jingyu Liu}{gl}
\icmlauthor{Huayi Tang}{gl}
\icmlauthor{Yong Liu}{gl,dp}
\end{icmlauthorlist}

\icmlaffiliation{gl}{Gaoling School of Artificial Intelligence, Renmin University of China, Beijing, China}
\icmlaffiliation{dp}{Beijing Key Laboratory of Big Data Management and Analysis Methods, Beijing, China}

\icmlcorrespondingauthor{Yong Liu}{liuyonggsai@ruc.edu.cn}

\icmlkeywords{Machine Learning, ICML}

\vskip 0.3in
]



\printAffiliationsAndNotice{} 

\begin{abstract}
   Graph Contrastive Learning (GCL) aims to learn node representations by aligning positive pairs and separating negative ones. However, few of researchers have focused on the inner law behind specific augmentations used in graph-based learning. What kind of augmentation will help downstream performance, how does contrastive learning actually influence downstream tasks, and why the magnitude of augmentation matters so much? This paper seeks to address these questions by establishing a connection between augmentation and downstream performance. Our findings reveal that GCL contributes to downstream tasks mainly by separating different classes rather than gathering nodes of the same class. So perfect alignment and augmentation overlap which draw all intra-class samples the same can not fully explain the success of contrastive learning. Therefore, in order to understand how augmentation aids the contrastive learning process, we conduct further investigations into the generalization, finding that perfect alignment that draw positive pair the same could help contrastive loss but is poisonous to generalization, as a result, perfect alignment may not lead to best downstream performance, so specifically designed augmentation is needed to achieve appropriate alignment performance and improve downstream accuracy. We further analyse the result by information theory and graph spectrum theory and propose two simple but effective methods to verify the theories. The two methods could be easily applied to various GCL algorithms and extensive experiments are conducted to prove its effectiveness. The code is available at \url{https://github.com/somebodyhh1/GRACEIS}

\end{abstract}

\section{Introduction}
    Graph Neural Networks (GNNs) have been successfully applied in various fields \cite{add_GNN_ego} such as recommendation systems \citep{GNN_for_recommendation}, drug discovery \citep{GNN_for_drug}, and traffic analysis \citep{GNN_for_traffic}, etc \cite{add_GNN1,add_GNN2,gnn_huayi}. However, most GNNs require labeled data for training, which may not always be available. To address this issue, Graph Contrastive Learning (GCL), which does not rely on labels, has gained popularity as a new approach to graph representation learning \citep{DGI,GCL_GraphCL}.

    GCL often generates new graph views through data augmentation \citep{CL_simCLR,GRACE,add_aug1,add_aug2}. GCL considers nodes augmented from the same as positive samples and others as negative samples. Subsequently, the model try to maximize similarity between positive samples and minimize similarity between negative ones \citep{CL_alignment_divergence,GCL_multi_view} to learn a better representation. So, data augmentation plays a vital role in graph contrastive learning, and data augmentation can be categorized into three types \citep{GCL_survey_rule_learn}: random augmentation \citep{DGI,GRACE}, rule-based augmentation \citep{GCL_GCA,GCL_saliency,GCL_specturm_NCE}, and learning-based augmentation \citep{GCL_AD_GCL,GCL_learn}. For instance, GRACE \citep{GRACE} randomly masks node attributes and edges in graph data to obtain augmented graphs; GCA \citep{GCL_GCA} uses node degree to measure its importance and mask those unimportant with higher probability; And AD-GCL \citep{GCL_AD_GCL} uses a model to learn the best augmentation and remove irrelevant information as much as possible. However, most data augmentation algorithms are designed heuristically, and there is a lack of theoretical analysis on how these methods will influence the downstream performance.

    Some researchers have explored the generalization ability of contrastive learning \citep{CL_first,CL_alignment_divergence,CL_hwr}. They propose that contrastive learning works by gathering positive pairs and separating negative samples uniformly. \citet{chaos} argues that perfect alignment and uniformity alone cannot guarantee optimal performance. They propose that through stronger augmentation, there will be support overlap between different intra-class samples, which is called augmentation overlap \citep{CL_inductive_bias,CL_hwr}. The augmentation overlap between two nodes mean that their corresponding augmented node could be the same, in this way aligning the anchor node with the the augmented node could also align two anchor nodes. Thus, the alignment of positive samples will also cluster all the intra-class samples together. And due to the limited inter-class overlap, inter-class nodes will not be gathered. However, \citet{CL_inductive_bias} points out that augmentation overlap may be relatively rare despite the excellent performance of contrastive learning methods. Hence, chances are that the success of contrastive learning cannot be solely attributed to alignment and augmentation overlap. It is of vital importance to figure out how augmentation works in the contrastive learning process, why the magnitude of augmentation matters so much and how to perform better augmentation. As data augmentation on graphs could be more customized and the magnitude of augmentation can be clearly represented by the number of modified edges/nodes \citep{GCL_GraphCL}, we mainly study the augmentation on graphs.

    In this paper, we provide a new understanding of Graph Contrastive Learning and use a theoretical approach to analyze the impact of augmentation on contrastive learning process. We find that with a stronger augmentation, the model is performing better mainly because of inter-class separating rather than intra-class gathering brought by augmentation overlap. This aligns with the finding that augmentation overlap is actually quite rare in contrastive learning \citep{CL_inductive_bias}. Also, \cite{chaos} proposes that a stronger augmentation could help because of more augmentation overlap, then more intra-class nodes are gathered due to alignment. However, stronger augmentation naturally conflicts with better alignment, so does stronger augmentation helps intra-class gathering remains questionable. Moreover, stronger augmentation leads to better performance while the alignment is weaken, so we also question that does perfect alignment actually helps contrastive learning?
    
    To further analyze the phenomena, we formulate a relationship between downstream accuracy, contrastive learning loss, and alignment performance, find that weak alignment performance caused by stronger augmentation can benefit the generalization. This explains why stronger augmentation will lead to better performance and reveals that perfect alignment is not the key to success, it may help to decrease contrastive loss, but also enlarge the gap between contrastive learning and downstream task, so specifically designed augmentation strategy is needed to achieve appropriate alignment and get the best downstream accuracy. This is why augmentation matters so much in contrastive learning.

    Then, aiming to achieve better downstream accuracy, we need to figure out how to perform augmentation to achieve a better balance between contrastive loss and generalization. Therefore, we further analyze the contrastive process through information theory and graph spectrum theory. From the information theory perspective, we find augmentation should be stronger while keeping enough information, which is widely adopted explicitly or implicitly by designed algorithms \citep{GCL_GCA,GRACE,GCL_AD_GCL}. From the graph spectrum theory perspective, we analyze how the graph spectrum will affect the contrastive loss and generalization \citep{GCL_specturm_NCE}, finding that non-smooth spectral distribution will have a negative impact on generalization. Then we propose two methods based on the theories to verify our findings.

    Our main contributions are as listed follows.    
    (1) We reveal that when stronger augmentation is applied, contrastive learning benefits from inter-class separating more than intra-class gathering, and better alignment may not be helping as it conflicts with stronger augmentation. (2) We establish the relationship between downstream performance, contrastive learning loss, and alignment performance. Finds that better alignment would weaken the generalization, showing that why stronger augmentation helps, then we analyze the result from information theory and graph spectrum theory. (3) Based on the proposed theoretical results, we provide two simple but effective algorithms to verify the correctness of the theory. We also show that these algorithms can be extended to various contrastive learning methods to enhance their performance. (4) Extensive experiments are conducted on different contrastive learning algorithms and datasets using our proposed methods to demonstrate its effectiveness and verify our theory.

\section{Augmentation and Generalization}

\subsection{Preliminaries}
    A graph can be represented as $\gG=(\gV,\gE)$, where $\gV$ is the set of $N$ nodes and $\gE \subseteq \gV \times \gV$ represents the edge set. The feature matrix and the adjacency matrix are denoted as $\mX \in \R^{N\times F}$ and $\mA \in \{0,1\}^{N\times N}$, where $F$ is the dimension of input feature, $\vx_i \in \R^F$ is the feature of node $v_i$ and $\mA_{i,j}=1$ iff $(v_i,v_j)\in \gE$. The node degree matrix $\mD=\mathrm{diag}(d_1,d_2,...,d_N)$, where $d_i$ is the degree of node $v_i$.

    In contrastive learning, data augmentation is used to create new graphs $\gG^1, \gG^2 \in \sG^{\text{aug}}$, and the corresponding nodes, edges, and adjacency matrices are denoted as $\gV^1, \gE^1, \mA^1, \gV^2, \gE^2, \mA^2$. In the following of the paper, $v$ is used to represent all nodes including the original nodes and the augmented nodes; $v_i^+$ is used to represent the augmented nodes including both $v_i^1$ and $v_i^2$; $v_i^0$ represents the original nodes only.

    Nodes augmented from the same one, such as $(v_i^1,v_i^2)$, are considered as positive pairs, while others are considered as negative pairs. It is worth noting that a negative pair could come from the same graph, for node $v_i^1$, its negative pair could be $v_i^-\in \{ v_j^{+} | j \neq i\}$. Graph Contrastive Learning (GCL) is a method to learn an encoder that draws the embeddings of positive pairs similar and makes negative ones dissimilar \citep{CL_simCLR,CL_alignment_divergence}. The encoder calculates the embedding of node $v_i$ by $f(v_i)$, and we assume that $||f(v_i)||=1$. 

\subsection{How Does Augmentation Affect Downstream Performance}

    Previous work \citep{CL_alignment_divergence} proposes that effective contrastive learning should satisfy alignment and uniformity, meaning that positive samples should have similar embeddings, \ie $f(v_i^1) \approx f(v_i^2)$, and features should be uniformly distributed in the unit hypersphere. However, \citet{chaos} pointed out that perfect alignment and uniformity does not ensure great performance. For instance, when $\{f(v_i^0)\}_{i=1}^N$ are uniformly distributed and $f(v_i^0)=f(v_i^+)$, there is a chance that the model may converge to a trivial solution that only projects very similar features to the same embedding, and projects other features randomly, then it will perform random classification in downstream tasks although it achieves perfect alignment and uniformity.

    \citet{chaos} argues that perfect alignment and intra-class augmentation overlap would be the best solution. The augmentation overlap means support overlap between different intra-class samples, and stronger augmentation is likely to bring more augmentation overlap. If two intra-class samples have augmentation overlap, then the best solution is projecting the two samples and their augmentation to the same embedding, which is called perfect alignment. For example, if two different nodes $v_i^0$, $v_j^0$ get the same augmentation $v^+$, then the best solution to contrative learning is $f(v_i^0)=f(v^+)=f(v_j^0)$. As the intra-class nodes are naturally closer, augmentation overlap often occurs between intra-class nodes, so perfect alignment and augmentation overlap could help intra-class gathering.

    However, \citet{CL_inductive_bias} proposes that augmentation overlap is actually quite rare in practice, even with strong augmentation. Also augmentation overlap requires for strong augmentation, which makes alignment harder and conflicts with perfect alignment, so the success of contrastive learning may not be contributed to intra-class gathering brought by augmentation overlap. Therefore, it is important to understand how is contrastive learning working, and why stronger augmentation helps. More related work are introduced in Appendix \ref{appendix:related_work}.
    
    To begin with, we give an assumption on the label consistency between positive samples, which means the class label does not change after augmentation.
    \begin{assumption}[View Invariance] \label{assumption:view_invariance}
        For node $v_i^0$, the corresponding augmentation nodes $v_i^+$ get consistent labels, \ie we assume the labels are deterministic (one-hot) and consistent: $p(y|v_i^0)=p(y|v_i^+)$.
    \end{assumption}
    This assumption is widely adopted \citep{CL_first,chaos,CL_inductive_bias} and reasonable. If the augmentation still keeps the basic structure and most of feature information is kept, the class label would not likely to change. Else if the augmentation destroys basic label information, the model tends to learn a trivial solution, so it is meaningless and we do not discuss the situation. The graph data keeps great label consistency under strong augmentation as discussed in Appendix \ref{appendix:delta_and_KL}.

    To further understand how is data augmentation working in contrastive learning, we use graph edit distance (GED) to denote the magnitude of augmentation, \citet{GCL_GED} proposes that all allowable augmentations can be expressed using GED which is defined as minimum cost of graph edition (node insertion, node deletion, edge deletion, feature transformation) transforming graph $\gG^0$ to $\gG^+$. So a stronger augmentation could be defined as augmentation with a larger graph edit distance.
    \begin{figure*}
        \centering
        \includegraphics[width=1\textwidth]{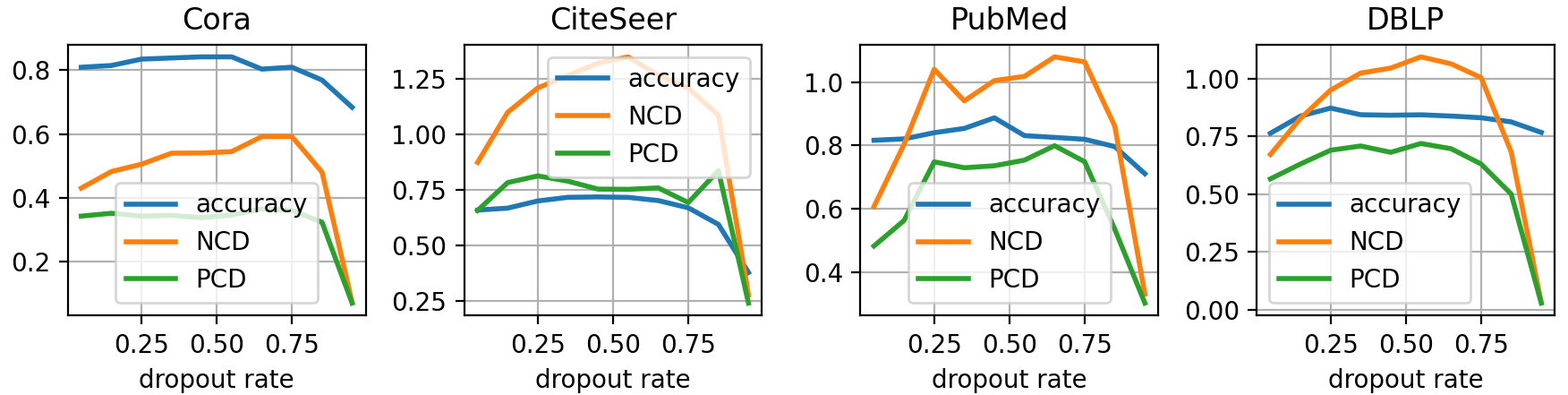}
        \caption{PCD means positive center distance ($\E_{p(v_y^{0}|y)}||f(v_y^{0})-\mu_y||$), NCD means negative center distance ($\E_{p(v_y^{0}|y)}||f(v_y^{0})-\mu_{y^-}||$) and accuracy is the downstream performance. X-axis stands for dropout rate of both edge and feature.}
        \label{fig:simi_simy}
    \end{figure*}
    \begin{assumption}[Augmentation Distance and Augmentation]\label{assumption augmentation distance and augmentation}
        While Assumption \ref{assumption:view_invariance} holds \ie $p(y|v_i^0)=p(y|v_i^+)$, as the augmentation getting stronger, the augmentation distance $\delta_{aug}^2=\E_{p(v_i^0,v_i^+)}||f(v_i^0)-f(v_i^+)||^2$ will increase, \ie $\delta_{aug} \propto \mathrm{GED}(\gG^0,\gG^+)$. $\mathrm{GED}(\gG^0,\gG^+)$ indicates the graph edit distance between $\gG^0$ and $\gG^+$.
    \end{assumption}
    This is a natural assumption that is likely to hold because a bigger difference in input will lead to a bigger difference in output. Also we can notice that $\delta_{aug}$ is actually the distance between the anchor node and the augmented node, so $\delta_{aug}$ could naturally represent the alignment performance, a smaller $\delta_{aug}$ means a better alignment. Then Assumption \ref{assumption augmentation distance and augmentation} means that stronger augmentation would lead to larger $\delta_{aug}$, \ie worse alignment. This phenomena is common in real practice as shown in Appendix \ref{appendix:delta_and_KL}.

    \begin{definition}
        The class center is calculated by the expectation of all nodes belongs to the class, \ie $\mu_y=\E_{p(v,y)}\left[f(v_y)\right]$. We use $\delta_{y^+}$ and $\delta_{y^-}$ to represent intra-class and inter-class divergence respectively, and
        \begin{equation}
            \begin{split}
                \delta_{y^+}^2&=\E_{p(y,i,j)}||f(v^0_{y,i})-f(v^0_{y,j})||^2,\\
                \delta_{y^-}^2&=\E_{p(y,y-,i,j)}||f(v^0_{y,i})-f(v^0_{y^-,j})||^2,
            \end{split}
            \nonumber
        \end{equation}
        where $y^-$ stands for a class different from $y$.
    \end{definition}

    Note that we calculate the class center $\mu_y$ by averaging nodes from both original view and augmented views. As the augmentation on graphs are highly biased \citep{zhang2022costa}, \ie the mean of augmented nodes are different from the original node, so the class center tends to be different. Also contrastive learning actually learns embedding on the augmented view, so the class gathering result is largely affected by the augmentation, so it is more appropriate to include the augmented nodes when calculate the class center.

    With the assumptions, we can get the theorem below:
    \begin{theorem}[Augmentation and Classification] \label{theorem:aug_cla}
        If Assumption \ref{assumption:view_invariance} holds, we know that:
        \begin{align}
            \E_{p(v_y^{0}|y)}||f(v_y^{0})-\mu_y|| \leq \delta_{y^+}+\frac{2}{3}\delta_{aug} \label{equation:positive center distance},\\
            \E_{p(v_y^{0}|y)}||f(v_y^{0})-\mu_{y^-}|| \leq \delta_{y^-}+\frac{2}{3}\delta_{aug} \label{equation:negative center distance},
        \end{align}
    \end{theorem}

    The proof can be found in Appendix \ref{proof:theorem augmentation and classification}. This shows that the distance between a node and the class center could be represented by the augmentation distance $\delta_{aug}$ and the inter-class/intra-class divergence $\delta_{y^-}$, $\delta_{y^+}$. We then use positive and negative center distance to represent $\E_{p(v_y^{0}|y)}||f(v_y^{0})-\mu_y||$ and $\E_{p(v_y^{0}|y)}||f(v_y^{0})-\mu_{y^-}||$, respectively.
    
    As we assumed in Assumption \ref{assumption augmentation distance and augmentation}, when the augmentation becomes stronger, augmentation distance \ie $\delta_{aug}$ would increase. Also we notice that both positive and negative center distance are positively related to the magnitude of augmentation $\delta_{aug}$. Therefore, stronger augmentation separates both inter-class and intra-class nodes, \ie it helps inter-class separating and hinders intra-class gathering. But the downstream performance tends to be better with stronger augmentation \citep{chaos,GRACE}, so the performance gain may be brought by inter-class separating more than intra-class gathering. 
    
    The experiment shown in Figure \ref{fig:simi_simy} confirms our suspicion. We use dropout on edges and features to perform augmentation, and the dropout rate naturally represents the magnitude of augmentation \ie graph edit distance. We present the positive/negative center distance and downstream accuracy to show the changing tendency. Figure \ref{fig:simi_simy} shows that initially, as the dropout rate increases, positive center distance is not decreasing, but downstream performance could be enhanced as negative center distance increases sharply. So the better performance correlates to inter-class separating, and the intra-class nodes may not be gathered.

    We show the results on more datasets including shopping graph, graph with heterophily and coauthor network in Appendix \ref{appendix:more_PCS}. From those experiments, we can conclude that contrastive learning mainly contributes to downstream tasks by separating nodes of different classes (increasing negative center distance) rather than gathering nodes of the same class (non-decreasing positive center distance). This explains why contrastive learning can achieve satisfying performance with limited augmentation overlap and relatively weak alignment \citep{CL_inductive_bias}.

    We can also understand the phenomena intuitively, The InfoNCE loss $\gL_{\mathrm{NCE}}$ can be written as below: 
    $$\gL_{\mathrm{NCE}}=\E_{p(v_i^1,v_i^2)}\E_{p({v_i^-})}\left[-\log\frac{\exp(f(v_i^1)^Tf(v_i^2))}{\sum\exp(f(v_i^1)^Tf(v^-_i))}\right].$$
    The numerator stands for positive pair similarity, so stronger augmentation would make positive pair dissimilar and the numerator is harder to maximize. Then GCL would pay more attention to the minimize the denominator as shown in Appendix \ref{appendix:change on positive/negative pair similarity}. Minimizing the denominator is actually separating negative samples, and separating negative samples could effectively separate inter-class nodes as most negative samples are from the different classes. In contrast, with stronger augmentation augmentation overlap is still quite rare and positive pair are harder to be aligned, so intra-class nodes are hard to be gathered while the existence of intra-class negative nodes further weaken intra-class gathering. As a result intra-class nodes may not gather closer during contrastive learning. Also we can observe from Figure \ref{fig:simi_simy} that when we drop too much edges/features, downstream performance decreases sharply, and both positive and negative center similarity increases as too much information is lost and the basic assumption $p(y|v_i^0)=p(y|v_i^+)$ does not hold, then a trivial solution is learned.
    
\subsection{Augmentation and Generalization}

    Although GCL with a stronger augmentation may help to improve downstream performance, why it works stays unclear. We need to figure out the relationship between augmentation distance, contrastive loss and downstream performance to further guide algorithm design. We first define the mean cross-entropy (CE) loss below, and use it to represent downstream performance.

    \begin{definition}[Mean CE loss]\label{definition: mean_CE_loss}
        For an encoder $f$ and downstream labels $y\in \left[1,K\right]$, we use the mean $\mathrm{CE}$ loss $\hat{\gL}_{\mathrm{CE}}=\E_{p(v^0,y)}\left[-\log \frac{\exp(f(v^0)^T\mu_y)}{\sum_{j=1}^K \exp(f(v^0)^T\mu_j)}\right]$ to evaluate downstream performance, where $\mu_j=\E_{p(v|y=j)}\left[f(v)\right]$.
    \end{definition}
    It is easy to see that mean CE loss could indicate downstream performance as it requires nodes similar to their respective class center, and different from others class centers. Also it is an upper bound of CE loss $\gL_{\mathrm{CE}}=\E_{p(v^0,y)}\left[-\log \frac{\exp(f(v^0)^T\omega_y}{\sum_{i=1}^K\exp(f(v^0)^T\omega_{i})}\right]$, where $\omega$ is parameter to train a linear classifier $g(z)=Wz$, $W=[\omega_1,\omega_2,...,\omega_k]$. \citet{CL_first} showed that the mean classifier could achieve comparable performance to learned weights, so we analyze the mean CE loss instead of the CE loss in this paper.

    \begin{theorem}[Generalization and Augmentation Distance] \label{theorem:generalization}
        If Assumption \ref{assumption:view_invariance} holds, and $\mathrm{ReLU}$ is applied as activation, then the relationship between downstream performance and $\mathrm{InfoNCE}$ loss could be represented as:
        \begin{align*}
            \hat{\gL}_{\mathrm{CE}} &\geq \gL_{\mathrm{NCE}}-3\delta_{aug}^2-2\delta_{aug}-\log\frac{M}{K}-\frac{1}{2}\var(f(v^+)|y) \nonumber\\
            &\hspace{1.5em}-\sqrt{\var(f(v^0)|y)}-e\var(\mu_y)-O(M^{-\frac{1}{2}}),\\
        \end{align*}
        where $M$ is number of negative samples\footnote{the generalization are correlated with $-\log M-O(M^{-\frac{1}{2}})$, which is decreasing when $M$ increases and $M$ is large, so the theorem encourages large negative samples.}, $K$ is number of classes.
    \end{theorem}
    
    The proof can be found in Appendix \ref{proof: theorem generalization}. Theorem \ref{theorem:generalization} gives a lower bound on the mean CE loss, we find that when we perform stronger augmentation, the lower bound would be smaller. The smaller lower bound does not enforce a better performance, but it shows a potential better solution. When the lower bound becomes smaller, the best solution is better so the model potentially performs better. For example, if there exists two models with $\hat{\gL}_{\mathrm{CE}} \geq 0.7$ and $\hat{\gL}_{\mathrm{CE}} \geq 0.3$ respectively, the latter one would be prefered as it is more likely to perform better, and the former one can never achieve performance better than 0.7. The upper bound instead shows the worst case, smaller upper bound means that the model could perform better at the worst case. From experimental results shown in Appendix \ref{appendix:more_PCS}, we can observe that the downstream performance tends to be better with stronger augmentation which corresponds to the decreasing lower bound, so the model is powerful enough to be close to the lower bound. Therefore, a smaller lower bound could lead to better performance.
    
    Theorem \ref{theorem:generalization} suggests a gap between $\hat{\gL}_{\mathrm{CE}}$ and $\gL_{\mathrm{NCE}}$, meaning that the encoders that minimize $\gL_{\mathrm{NCE}}$ may not yield optimal performance on downstream tasks. Furthermore, it suggests that a higher augmentation distance $\delta_{aug}$ would make the bound smaller and enhance generalization, improve performance on downstream tasks. This aligns with previous finding that a stronger augmentation helps downstream performance. Also Inequality (\ref{equation:positive center distance}) demonstrates that the positive center distance is positively related to $\delta_{aug}$, so better generalization correlates with higher positive center distance. This aligns with the experiments before that better downstream performance may come with a high positive center distance.
    
    Theorem \ref{theorem:generalization} also highlights the significance of augmentation magnitude in graph contrastive learning algorithms like GRACE \citep{GRACE}. A weak augmentation leads to better alignment but also a weak generalization, InfoNCE loss might be relatively low but the downstream performance could be terrible \citep{CL_inductive_bias}. When augmentation gets stronger, although perfect alignment cannot be achieved, it promotes better generalization and potentially leads to improved downstream performance. And when the augmentation is too strong, minimizing the InfoNCE loss becomes challenging \citep{GCL_invariant}, leading to poorer downstream performance. Therefore, it is crucial to determine the magnitude of augmentation and how to perform augmentation as it directly affects contrastive performance and generalization.

\section{Finding Better Augmentation}

    Previous sections have revealed that perfect alignment, may not help downstream performance. Instead a stronger augmentation that leads to larger $\delta_{aug}$ will benefit generalization while weakening contrastive learning process. Therefore, we need to find out how to perform augmentation to strike a better balance between augmentation distance and contrastive loss, leading to better downstream performance.

\subsection{Information Theory Perspective}\label{section:information theory perspective}

    Due to the inherent connection between contrastive learning and information theory, we try to analyse it through information perspective. As shown by \citet{NCE_MI}, $\gL_{\mathrm{NCE}}$ is a lower bound of mutual information. And, $\var(f(v^0)|y)$, $\var(f(v^+|y))$ and $\var(\mu_y)$ can be represented by inherent properties of the graph and the augmentation distance $\delta_{aug}$. Thus, we can understand the process through information and augmentation, we can reformulate Theorem \ref{theorem:generalization} as follows:

    \begin{corollary}[CE with Mutual Information]\label{corollary:CE_MI}
        If Assumption \ref{assumption:view_invariance} holds, the relationship between downstream performance, mutual information between views and augmentation distance could be represented as:
        \begin{equation}
            \hat{\gL}_{\mathrm{CE}}\geq \log(K)-I(v^1;v^2)-g(\delta_{aug})-O(M^{-\frac{1}{2}}),
            \nonumber
        \end{equation}
        where $I(v^1;v^2)$ stands for the mutual information between $v^1$ and $v^2$, $g(\delta_{aug})$ is monotonically increasing with $\delta_{aug}$, and is defined in Appendix \ref{proof:corollary CE_MI}.
    \end{corollary}
    
    The proof can be found in Appendix \ref{proof:corollary CE_MI}. Corollary \ref{corollary:CE_MI} suggests that the best augmentation would be one that maximize the mutual information and augmentation distance. \citet{CL_good_view} propose that a good augmentation should minimize $I(v^1;v^2)$ while preserve as much downstream related information as possible, \ie $I(v^1;y)=I(v^2;y)=I(v^0;y)$. However, downstream tasks is unknown while pretraining, so this is actually impossible to achieve. Our theory indicates that the augmentation should be strong while preserving as much information as possible, and the best augmentation should be the one satisfying InfoMin which means the augmentation gets rid of all useless information and keeps the downstream related ones. So InfoMin propose the ideal augmentation which can not be achieved, and we propose an actual target to train a better model.
    
    To verify our theory, we propose a simple but effective method. We first recognize important nodes, features and edges, then leave them unchanged during augmentation to increase mutual information. Then for those unimportant ones, we should perform stronger augmentation to increase the augmentation distance.
    
    We utilize gradients to identify which feature of node $v$ is relatively important and carries more information. We calculate the importance of feature by averaging the feature importance across all nodes, the importance of node $v$ could be calculated by simply averaging the importance of its features, and then use the average of the two endpoints to represent the importance of an edge:
    \begin{equation}
        \begin{split}
            &\alpha_{v,p}=\frac{\partial \gL_{\mathrm{NCE}}}{\partial x_{v,p}},\quad \alpha_{p}=\mathrm{ReLU}\left(\frac{1}{|V'|}\sum_{v}\alpha_{v,p}\right),\\
            &\alpha_{v}=\mathrm{ReLU}\left(\frac{1}{|P'|}\sum_{p}\alpha_{v,p}\right), \quad \alpha_{e_{i,j}}=\left(\alpha_{v_i}+\alpha_{v_j}\right)/2,
        \end{split}
        \nonumber
    \end{equation}
    where $\alpha_{v,p}$ is importance of the $p^{th}$ feature of node $v$, $\alpha_p$ is the importance of $p^{th}$ feature, $\alpha_{v}$ iss importance of node $v$, and $\alpha_{e_{i,j}}$ means the importance of edge $(v_i,v_j)$.
    
    For those edges/features with high importance, we should keep them steady and do no modification during augmentation. For those with relatively low importance, we can freely mask those edges/features, but we should make sure that the number of masked edges/features is greater than the number of kept ones to prevent $\delta_{aug}$ from decreasing. The process can be described by the following equation:
    \begin{equation}
        \tilde{\mA}=\mA \ast (\mM_e \vee \mS_e \wedge \mD_e), \quad 
        \tilde{\mF}=\mF \ast (\mM_f \vee \mS_f \wedge \mD_f),
        \nonumber
    \end{equation}

    where $\ast$ is hadamard product, $\vee$ stands for logical OR, $\wedge$ stands for logical AND. $\mM_e$, $\mM_f$ represent the random mask matrix, which could be generated using any masking method, $\mS_e$, $\mS_f$ are the importance based retain matrix, it tells which edge/feature is of high importance and should be retained. For the top $\xi$ important edges/features, we set $\mS_e$, $\mS_f$ to $1$ with a probability of 50\% and to $0$ otherwise. $\mD_e$, $\mD_f$ show those edges/features should be deleted to increase $\delta_{aug}$, for the least 2$\xi$ important edges/features, we also set $\mD_e$, $\mD_f$ to $0$ with a probability of 50\% and to $1$ otherwise.

    This is a simple method, and the way to measure importance can be replaced by any other methods. It can be easily integrated into any other graph contrastive learning methods that require edge/feature augmentation. There are many details that could be optimized, such as how to choose which edges/features to delete and the number of deletions. However, since this algorithm is primarily intended for theoretical verification, we just randomly select edges to be deleted and set the number to be deleted as twice the number of edges kept.
    
    In fact, most graph contrastive learning methods follow a similar framework as discussed in Appendix \ref{appendix:spatial augmentation and MI}.

    \renewcommand\arraystretch{1}
\begin{table*}[htbp]
  \centering
      \caption{Quantitative results on node classification, algorithm+I stands for the algorithm with information augmentation, and algorithm+S stands for the algorithm with spectrum augmentation. We show the error bar in Figure \ref{fig:error_bar}} \label{table:results_of_algorithm}
  \resizebox{0.9\textwidth}{!}{
    \begin{tabular}{c|ccccccccc}
    \toprule
    \multicolumn{2}{c}{Methods} & Cora  & CiteSeer & PubMed & DBLP  & Photo & Computers & mean  & p-value \\
    \midrule
    \multirow{4}[2]{*}{\rotatebox{90}{Baseline}} & Supervised GCN & 83.31±0.07 & 69.81±0.98 & 85.36±0.09 & 81.26±0.01 & 93.28±0.03 & 88.11±0.14 & 81.93  & - \\
          & Supervised GAT & 83.83±0.30 & 70.31±0.65 & 84.04±0.40 & 81.92±0.03 & 93.17±0.05 & 86.82±0.09 & 81.57  &-  \\
          & GRACE+SpCo & 83.45±0.79 & 69.9±1.24 & OOM   & 83.61±0.14 & 91.56±0.19 & 83.37±0.38 & 81.01  &-  \\
          & GCS   & 83.39±0.54 & 68.73±1.68 & 84.92±0.19 & 83.38±0.37 & 90.15±0.24 & 86.54±0.26 & 81.97  &-  \\
    \midrule
    \multirow{9}[6]{*}{\rotatebox{90}{Unsupervised}} & GRACE & 82.52±0.75 & 70.44±1.49 & 84.97±0.17 & 84.01±0.34 & 91.17±0.15 & 86.36±0.32 & 83.25  &-  \\
          & GRACE+I & \textbf{83.78±1.08} & \textbf{72.89±0.97} & 84.97±0.14 & 84.80±0.17 & 91.64±0.21 & 87.57±0.53 & 84.28  & 0.155 \\
          & GRACE+S & 83.61±0.85 & 72.83±0.47 & \textbf{85.45±0.25} & \textbf{84.83±0.18} & \textbf{91.99±0.35} & \textbf{87.67±0.33} & \textbf{84.40 } & 0.003 \\
\cmidrule{2-10}          & GCA   & 83.74±0.79 & 71.09±1.29 & 85.38±0.20 & 83.99±0.21 & 91.67±0.38 & 86.77±0.31 & 83.77  &-  \\
          & GCA+I & \textbf{84.93±0.81} & \textbf{72.74±1.05} & \textbf{85.73±0.13} & \textbf{84.79±0.28} & 91.94±0.13 & 86.60±0.29 & \textbf{84.46 } & 0.089 \\
          & GCA+S & 84.51±0.89 & 72.38±0.86 & 85.35±0.09 & 84.49±0.24 & \textbf{92.02±0.34} & \textbf{86.97±0.40} & 84.29  & 0.147 \\
\cmidrule{2-10}          & AD GCL & 81.68±0.80 & 70.01±0.97 & 84.77±0.15 & 83.14±0.16 & 91.34±0.33 & 84.80±0.51 & 82.62  &-  \\
          & AD GCL+I & \textbf{83.46±1.06} & 71.06±0.91 & \textbf{85.52±0.33} & \textbf{84.76±0.09} & 91.71±0.78 & 86.02±0.53 & \textbf{83.76 } & 0.003 \\
          & AD GCL+S & 82.96±0.53 & \textbf{71.35±0.47} & 85.38±0.30 & 84.45±0.19 & \textbf{91.79±0.33} & \textbf{86.49±0.26} & 83.74  & 0.006 \\
    \bottomrule
    \end{tabular}%
    }
\end{table*}%

    \renewcommand\arraystretch{1}

\subsection{Graph Spectrum Perspective}
    In this section, we attempt to analyze InfoNCE loss and augmentation distance from graph spectrum perspective as graph and GNNs are deeply connected with spectrum theory. We start by representing them using the spectrum of the adjacency matrix $\mA$.
    \begin{theorem}[Theorem 1 of \citet{GCL_specturm_NCE} Restated]\label{theorem:spec_NCE}
        Given adjacency martix $\mA$ and the generated augmentation $\mA', \mA''$, the $i^{th}$ eigenvalues of $\mA'$ and $\mA''$ are $\lambda_i',\lambda_i''$, respectively. The following upper bound is established:
        \begin{equation}
            \begin{split}
                \gL_{\mathrm{NCE}} &\geq N\log N - (N+1)\sum_i\theta_i\lambda_i'\lambda_i'',\\
            \end{split}
            \label{equation:NCE_spectrum}
        \end{equation}
        where $\theta_i$ is the adaptive weight of the $i^{th}$ term, the detail of $\theta_i$ is discussed in $\mathrm{Appendix}$ $\mathrm{\ref{appendix:value of thetas}}$.
    \end{theorem}
    
    \begin{corollary}[Spectral Representation of $\delta_{aug}$]\label{corollary:spectrum_positive_pair_difference}
        If Assumption \ref{assumption:view_invariance} holds, and $\lambda_i',\lambda_i''$ are $i^{th}$ eigenvalues of $\mA'$ and $\mA''$, respectively, then:
        \begin{equation}
            2\delta_{aug} \geq \E_{p(v_i^1,v_i^2)}||f(v_i^1)-f(v_i^2)|| \geq \sqrt{2-\frac{2}{N}\sum_i\theta_i\lambda_i'\lambda_i''}. 
            \label{equation:spectrum_positive_pair_difference}
        \end{equation}
    \end{corollary}

    Theorem \ref{theorem:generalization} suggests that we should strive to make $\gL_{\mathrm{NCE}}$ small while increase $\delta_{aug}$, but they are kindly mutually exclusive. As shown in Theorem \ref{theorem:spec_NCE}, and Corallary \ref{corollary:spectrum_positive_pair_difference} proved in Appendix \ref{proof:corollary_spectrum_positive_pair_difference}, when $\theta_i$ is positive, a small $\gL_{\mathrm{NCE}}$ requires for large $|\lambda_i|$ while a large $\delta_{aug}$ requires for small $|\lambda_i|$, and it works exclusively too when $\theta_i$ is negative. As contrastive learning is trained to minimize $\gL_{\mathrm{NCE}}$, $\theta$s are going to increase as the training goes, so we can assume that $\theta$s will be positive, the detailed discussion and exact definition of $\theta$ can be found in Appendix \ref{appendix:value of thetas}. Therefore, to achieve a better trade-off, we should decrease $|\lambda_i|$ while keep InfoNCE loss also decreasing.
    \begin{figure*}[h]
        \centering
        \includegraphics[width=1\textwidth]{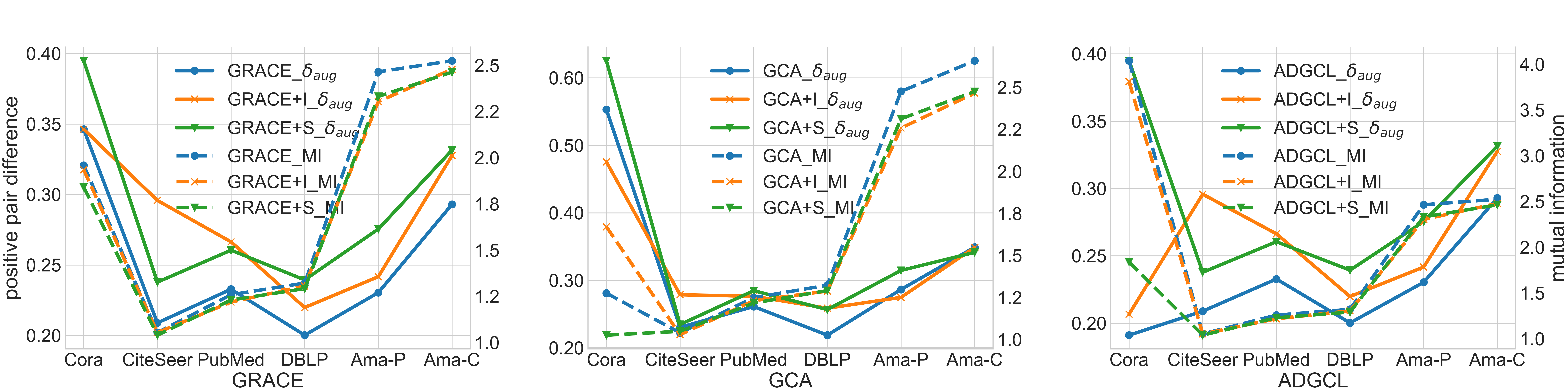}
        \caption{Augmentation distance and InfoNCE, GRACE+I stands for GRACE with information augmentation, and GRACE+S stands for GRACE with spectrum augmentation. GRACE+x$\_$MI means mutual information between two views after training, and GRACE+x$\_\delta_{aug}$ is augmentation distance caused by the method.}
        \label{fig:NCE_T_algorithm}
    \end{figure*}
    \begin{figure*}[h]
        \centering
        \includegraphics[width=0.9\textwidth]{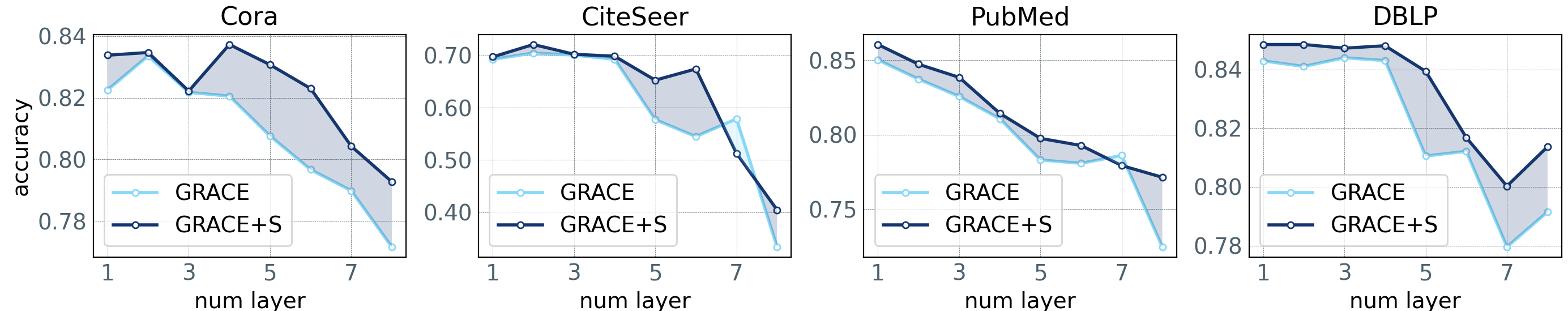}
        \caption{Accuracy on downstream tasks with different number of layers.}
        \label{fig:downstream layer}
    \end{figure*}
    In fact, reducing $|\lambda_i|$ actually reduces the positive $\lambda_i$ and increases the negative $\lambda_i$, which is trying to smoothen the graph spectrum and narrow the gap between the spectrum. As suggested by \citet{GRL_smooth_spectrum}, graph convolution operation with unsmooth spectrum results in signals correlated to the eigenvectors corresponding to larger magnitude eigenvalues and orthogonal to the eigenvectors corresponding to smaller magnitude eigenvalues. So with enough graph convolution operations, if $|\lambda_i| > |\lambda_j|$, then we can get the embedding $f(v)$ satisfying $\mathrm{sim}(f(v),e_i) \gg \mathrm{sim}(f(v),e_j)$ where $e_i$ denotes the eigenvector corresponding to $\lambda_i$, causing all representations similar to $e_i$. Therefore, an unsmooth spectrum may lead to similar representations and result in over-smooth. This can also be observed from Inequality (\ref{equation:spectrum_positive_pair_difference}), where a higher $|\lambda_i|$ draws $f(v_i^1)$ and $f(v_i^2)$ more similar.

    We now know that smoothing the graph spectrum can help with graph contrastive learning. The question is how to appropriately smooth the spectrum. We propose a simple method. As the training aims to minimize $\gL_{\mathrm{NCE}}$, the parameter $\theta_i$s are supposed to increase. Therefore, we can use $\theta_i$ as a symbol to show whether the model is correctly trained. When $\theta_i$ gradually increases, we can decrease $\lambda$ as needed. However, when $\theta_i$ starts to decrease, it is likely that the change on the spectrum is too drastic, and we should take a step back. The process could be described as follows:
    \begin{equation}
    \begin{split}
        \lambda_i&=\lambda_i+\mathrm{direction}_i*\lambda_i*\alpha,\\
        \mathrm{direction}_i&=
        \begin{cases}
            -1, & \text{$\mathrm{cur}(\theta_i)-\mathrm{pre}(\theta_i) \geq \epsilon$}\\
            1, & \text{$\mathrm{cur}(\theta_i)-\mathrm{pre}(\theta_i) \leq -\epsilon$}, \\
            0, & \text{otherwise}
        \end{cases}
    \end{split}
        \nonumber
    \end{equation}
    where $\alpha$ is a hyperparameter that determines how much we should decrease/increase $\lambda_i$. $\epsilon$ is used to determine whether $\theta_i$ is increasing, decreasing, or just staying steady. $\mathrm{cur}(\theta_i)$ and $\mathrm{pre}(\theta_i)$ represents the current and previous $\theta_i$.
    
    In this way, the contrastive training will increase $\theta_i$ and result in a lower $\gL_{\mathrm{NCE}}$, while we justify $\lambda_i$ to achieve a better augmentation distance, which promises a better generalization ability. Also some spectral augmentations implicitly decreases $|\lambda|$s as shown in Appendix \ref{appendix:spectral augmentation and lambda}.

\section{Experiments}

    In this section, we mainly evaluate the performance of the methods we proposed on six datasets: Cora, CiteSeer, PubMed, DBLP, Amazon-Photo and Amazon-Computer. We select 3 contrastive learning GNN, GRACE \citep{GRACE}, GCA \citep{GCL_GCA} and AD-GCL \citep{GCL_AD_GCL}, then we integrate those models with our proposed methods to verify its applicability and correctness of the theory. Details of datasets and baselines are shown in Appendix \ref{appendix:datasets_baselines}. The results are summarized in Table \ref{table:results_of_algorithm}. We further investigate the positive/negative center distance in Appendix \ref{appendix:center_distance}, the hyperparameter sensitivity is studied in Appendix \ref{appendix:hyperparameter_sensitivity}, and the change of $\theta$ and the spectrum while training is shown in Appendix \ref{appendix:changes on the spectrum}.

    From Table \ref{table:results_of_algorithm} shows that GRACE+I (GRACE with information augmentation) and GRACE+S (GRACE with spectrum augmentation) both improve the downstream performance. This improvement is significant for GRACE since it primarily performs random dropout, resulting in the loss of valuable information. But for GCA, the performance gain is relatively weak as GCA already drops the unimportant ones with a higher probability, allowing it to capture sufficient information. AD-GCL aggressively drops as much information as possible and some important ones are also dropped, so our methods help greatly. Overall, our methods improve the performance of original algorithm and helps downstream tasks, the p-value on the averaged performance  shown in Table \ref{table:results_of_algorithm} also prove that our method is effective. We further discuss the two different methods and combine then in Appendix \ref{appendix:IS_combine}. Also we conduct further discussion on some augmentation free methods in Appendix \ref{appendix:augmentation_free_methods}.
    
\subsection{Augmentation Distance}

    Figure \ref{fig:NCE_T_algorithm} shows that for all three algorithms, our augmentation methodologies can conduct stronger augmentation while preserving similar mutual information. In this way, our methods achieve higher augmentation distance while capturing similar information of the original view. So our methods achieve similar contrastive loss with better generalization, resulting in improved downstream performance. 
\subsection{Over-smooth}

    While reducing $|\lambda_i|$, we obtain a graph with smoother spectrum, and could relieve the over-smooth by preventing nodes being too similar with the eigenvector corresponding to the largest eigenvalue. This enables the application of relatively more complex models. We can verify this by simply stacking more layers. As shown in Figure \ref{fig:downstream layer}, if applied spectrum augmentation, the model tends to outperform the original algorithm especially with more layer, and the best performance may come with a larger number of layers, which indicates that more complicated models could be applied and our method successfully relieve over-smooth.

\section{Conclusion}
    In this paper, we study the impact of contrastive learning on downstream tasks and propose that perfect alignment does not necessarily lead to better performance. Instead, we find that a relatively large augmentation distance is more beneficial for generalization by enlarging the distance of inter-class nodes. We further study how the augmentation influences contrastive learning by information theory and the graph spectrum theory and propose two effective methods.

\section*{Impact Statement}
This work studies the theory and algorithm of Graph Contrastive Learning, which does not present any foreseeable societal consequence.

\section*{Acknowledgements}
We thank the anonymous reviewers for their valuable and constructive suggestions and comments. This work is supported by the Beijing Natural Science Foundation (No.4222029); the National Natural Science Foundation of China (N0.62076234); the National Key Research and Development Project (No.2022YFB2703102); the ``Intelligent Social Governance Interdisciplinary Platform, Major Innovation \& Planning Interdisciplinary Platform for the ``Double-First Class'' Initiative, Renmin University of China''; the Beijing Outstanding Young Scientist Program (NO.BJJWZYJH012019100020098); the Public Computing Cloud, Renmin University of China; the Fundamental Research Funds for the Central Universities, and the Research Funds of Renmin University of China (NO.2021030199), the Huawei-Renmin University joint program on Information Retrieval: the Unicom Innovation Ecological Cooperation Plan; the CCF-Huawei Populus Grove Fund.

\bibliography{reference}

\bibliographystyle{icml2024}

\newpage
\appendix
\onecolumn
\section{Theoretical Proof}
\subsection{Proof of Theorem \ref{theorem:aug_cla}}\label{proof:theorem augmentation and classification}
If we set $\delta_{y^+}^2=\E_{p(y,i,j)}||f(v^0_{y,i})-f(v^0_{y,j})||^2$, and $\delta_{y^+}^2=\E_{p(y,y',i,j)}||f(v^0_{y,i})-f(v^0_{y',j})||^2$. Then with Assumption \ref{assumption:view_invariance} and jensen inequality, we know that $\E_{p(v_i)}||f(v_i^0)-f(v_i^+)||^2 \leq \delta_{aug}^2$, $\E_{p(v_i)}||f(v_i^0)-f(v_i^+)|| \leq \delta_{aug}$ and $\E_{p(y,i,j)}||f(v^0_{y,i})-f(v^0_{y,j})||\leq \delta_{y^+}$, $\E_{p(y,y',i,j)}||f(v^0_{y,i})-f(v^0_{y',j})||\leq \delta_{y^-}$.
Therefore, we can get the inequality below:
\begin{align*}
    \E_{p(v_{y,i},v_{y,j}|y)}||f(v^+_{y,i})-f(v^0_{y,j})||^2
    &\leq \E_{p(v_{y,i},v_{y,j}|y)}||f(v^+_{y,i})-f(v^0_{y,i})||^2+\E_{p(v_{y,i},v_{y,j}|y)}||f(v^0_{y,i})-f(v^0_{y,j})||^2\\
    &\hspace{1.5em}+2\E_{p(v_{y,i},v_{y,j}|y)}||f(v^+_{y,i})-f(v^0_{y,j})||\cdot||f(v^0_{y,i})-f(v^0_{y,j})|| \\
    &\leq \delta_{aug}^2+\delta_{y^+}^2+2\delta_{aug}\delta_{y^+}\\
    &=(\delta_{aug}+\delta_{y^+})^2.
\end{align*}

As $\mu_y=\E_{p(v_y|y)}[f(v_y)]=\frac{1}{3}\E_{p(v_y^0|y)}f(v_y^0)+\frac{2}{3}\E_{p(v_y^+|y)}f(v_y^+)$, we know that,

\begin{equation}
    \begin{split}
        \E_{p(v_y^{0'}|y)}||f(v_y^{0'})-\mu_y||
        &=\E_{p(v_y^{0'}|y)}||f(v_y^{0'})-\frac{1}{3}\E_{p(v_y^0|y)}f(v_y^0)-\frac{2}{3}\E_{p(v_y^+|y)}f(v_y^+)||\\
        &=\E_{p(v_y^{0'}|y)} \left\|\frac{1}{3}\left(f(v_y^{0'})-\E_{p(v_y^0|y)}f(v_y^0)\right)+\frac{2}{3}\left(f(v_y^{0'})-\E_{p(v_y^+|y)}f(v_y^+)\right)\right\|\\
        &\leq \E_{p(v_y^{0'}|y)} \left[ \left\|\frac{1}{3}\left(f(v_y^{0'})-\E_{p(v_y^0|y)}f(v_y^0)\right)\right\|+\left \|\frac{2}{3} \left(f(v_y^{0'})-\E_{p(v_y^+|y)}f(v_y^+)\right)\right \|  \right] \\
        &\leq \E_{p(v_y^{0'}|y)} \E_{p(v_y^0|y)} \frac{1}{3}||(f(v_y^{0'})-f(v_y^0))||+\E_{p(v_y^{0'}|y)}\E_{p(v_y^+|y)} \frac{2}{3}||f(v_y^{0'})-f(v_y^+)||\\
        &\leq \frac{1}{3} \delta_{y^+}+\frac{2}{3} (\delta_{aug}+\delta_{y^+})\\
        &=\delta_{y^+}+\delta_{aug}
    \end{split}
    \nonumber
\end{equation}
Similarly, we know that $\E_{p(v_y^{0'}|y)}||f(v_y^{0'})-\mu_{y^-}||\leq \delta_{y^-}+\delta_{aug}$

Next we prove a bound for $\E_{p(v_y^{0},y^-|y)}f(v_y^0)^T\mu_{y^-}$ for other use.
As $\mu_y=\E_{p(v_y|y)}[f(v_y)]=\frac{1}{3}\E_{p(v_y^0|y)}f(v_y^0)+\frac{2}{3}\E_{p(v_y^+|y)}f(v_y^+)$, we know that,
\begin{align*}
    \E_{p(v_y^{0'}|y)}f(v_y^{0'})^T\mu_y
    &=\E_{p(v_y^{0'}|y)}f(v_y^{0'})^T(\frac{1}{3}\E_{p(v_y^0|y)}f(v_y^0)+\frac{2}{3}\E_{p(v_y^+|y)}f(v_y^0))\\
    &=\frac{1}{3}\E_{p(v_y^{0'}|y)}\E_{p(v_y^{0}|y)}f(v^{0'}_y)^Tf(v_y^0)+\frac{2}{3}\E_{p(v_y^{0'}|y)}\E_{p(v_y^{+}|y)}f(v^{0'}_y)^Tf(v_y^+).
\end{align*}

assume that $\E_{p(a,b)}||a-b||^2\leq c^2$, $||a||=||b||=1$, then
\begin{align*}
    &\E_{p(a,b)}(a^T-b^T)(a-b)\leq c^2 \\
    &\E_{p(a,b)} [a^Ta-a^Tb-b^Ta+b^Tb]\leq c^2 \\
    &\E_{p(a,b)} [2-2a^Tb]\leq c^2 \\
    &\E_{p(a,b)} a^Tb\geq\frac{2-c^2}{2}=1-\frac{c^2}{2}.\\
\end{align*}

As we already know that $\E_{p(y,y',i,j)}||f(v^0_{y,i})-f(v^0_{y',j})||^2 \leq \delta_{y^+}^2$ and $\E_{p(v_{y,i},v_{y,j}|y)}||f(v^+_{y,i})-f(v^0_{y,j})||^2 \leq (\delta_{aug}+\delta_{y^+})^2$. So $\E_{p(v_y^{0'}|y)}\E_{p(v_y^{0}|y)}f(v^{0'}_y)^Tf(v_y^0) \geq 1-\frac{\delta_{y^+}^2}{2}$ and $\E_{p(v_y^{0'}|y)}\E_{p(v_y^{+}|y)}f(v^{0'}_y)^Tf(v_y^+) \geq 1-\frac{(\delta_{aug}+\delta_{y^+})^2}{2}$.

Then, we can calculate $\E_{p(v_y^{0'}|y)}f(v_y^{0'})^T\mu_y$ as below:
\begin{equation}
    \begin{split}
        \E_{p(v_y^{0'}|y)}f(v_y^{0'})^T\mu_y
        &=\frac{1}{3}\E_{p(v_y^{0'}|y)}\E_{p(v_y^{0}|y)}f(v^{0'}_y)^Tf(v_y^0)+\frac{2}{3}\E_{p(v_y^{0'}|y)}\E_{p(v_y^{+}|y)}f(v^{0'}_y)^Tf(v_y^+)\\
        &\geq 1-\frac{\delta_{aug}^2}{3}-\frac{2\delta_{aug}\delta_{y^+}}{3}-\frac{\delta_{y^+}^2}{2}.
    \end{split}
\end{equation}

Similarly, we know that $\E_{p(v_y^{0},y^-|y)}f(v_y^0)^T\mu_{y^-}\geq 1-\frac{\delta_{aug}^2}{3}-\frac{2\delta_{aug}\delta_{y^-}}{3}-\frac{\delta_{y^-}^2}{2}$.

\subsection{Proof of Theorem \ref{theorem:generalization}} \label{proof: theorem generalization}
$$
\hat{\gL}_{\mathrm{CE}}=\underbrace{-\E_{p(v_i^0,y)}f(v_i^0)^T\mu_y}_{\Lambda_1}+\underbrace{\E_{p(v_i^0)}\log\sum_{i=j}^K\exp(f(v_i^0)^T\mu_j)}_{\Lambda_2}.
$$
\begin{align*}
    \Lambda_1
    &=-\E_{p(v_i^0,y)}f(v_i^0)^T\mu_y \\
    &=-\E_{p(v_i^0,y)}\left[f(v_i^0)^Tf(v_i^+)+f(v_i^0)^T(\mu_y-f(v_i^+))\right] \\
    &\overset{(a)}{\geq}-\E_{p(v_i^0,v_i^+,y)}f(v_i^0)^Tf(v_i^+)-\E_{p(v_i^+,y)}||f(v_i^+)-\mu_y|| \\
    &\geq-\E_{p(v_i^0,v_i^+,y)}f(v_i^0)^Tf(v_i^+)-\E_{p(v_i^0,v_i^+,y)}||f(v_y^+)-f(v_y^0)||-\E_{p(v_i^0,v_i^+,y)}||f(v^0_y)-\mu_y|| \\
    &\overset{(b)}{\geq}-\E_{p(v_i^0,v_i^+,y)}f(v_i^1)^Tf(v_i^2)-3\delta_{aug}^2-\delta_{aug}-\E_{p(v_i^0,v_i^+,y)}||f(v^0_y)-\mu_y||.
\end{align*}
(a) $f(v_i^0)^T(\mu_y-f(v_i^+))\leq (\frac{\mu_y-f(v_i^+)}{||\mu_y-f(v_i^+)||})^T(\mu_y-f(v_i^+))=||\mu_y-f(v_i^+)||$.

(b) $\E_{p(v_i^0,v_i^+)}||f(v_i^0)-f(v_i^+)||^2\leq \delta_{aug}^2$, then:
\begin{align*}
        \delta_{aug}^2
        &\geq \E_{p(v_i^0,v_i^1)}(f(v_i^0)-f(v_i^1))^T\cdot(f(v_i^0)-f(v_i^1)) \\
        &=\E_{p(v_i^0,v_i^1,v_i^2)}(f(v_i^0)-f(v_i^1))^T \cdot (f(v_i^0)-f(v_i^1)+f(v_i^2)-f(v_i^2)) \\
        &=\E_{p(v_i^0,v_i^1,v_i^2)} f(v_i^0)^Tf(v_i^0)-f(v_i^0)^Tf(v_i^1)+f(v_i^0)^Tf(v_i^2)-f(v_i^0)^Tf(v_i^2) \\
        &\hspace{5em} -f(v_i^1)^Tf(v_i^0)+f(v_i^1)^Tf(v_i^1)-f(v_i^1)^Tf(v_i^2)+f(v_i^1)^Tf(v_i^2) \\
        &=2+\E_{p(v_i^0,v_i^1,v_i^2)} \left[-2f(v_i^0)^Tf(v_i^1)+f(v_i^0)^Tf(v_i^2)-f(v_i^0)^Tf(v_i^2)-f(v_i^1)^Tf(v_i^2)+f(v_i^1)^Tf(v_i^2)\right] \\
        &\overset{(c)}{\geq}2-2+\E_{p(v_i^0,v_i^1,v_i^2)} \left[f(v_i^0)^Tf(v_i^2)-1-f(v_i^1)^Tf(v_i^2)+1-2\delta_{aug}^2\right] \\
        &=\E_{p(v_i^0,v_i^1,v_i^2)} \left[f(v_i^0)^Tf(v_i^2)- f(v_i^1)^Tf(v_i^2)\right]-2\delta_{aug}^2. \\
\end{align*}

So, we can get the relationship between $\E_{p(v_i^0,v_i^1,v_i^2)} f(v_i^0)^Tf(v_i^2)$ and $\E_{p(v_i^0,v_i^1,v_i^2)} f(v_i^1)^Tf(v_i^2)-2\delta_{aug}^2$ as below:

\begin{gather*}
    \delta_{aug}^2\geq \E_{p(v_i^0,v_i^1,v_i^2)} f(v_i^0)^Tf(v_i^2)-\E_{p(v_i^0,v_i^1,v_i^2)} f(v_i^1)^Tf(v_i^2)-2\delta_{aug}^2, \\
    \E_{p(v_i^0,v_i^1,v_i^2)} f(v_i^0)^Tf(v_i^2)\leq \E_{p(v_i^0,v_i^1,v_i^2)} f(v_i^1)^Tf(v_i^2)+3\delta_{aug}^2.\\
\end{gather*}
As $v_i^2$ is an augmented node, we can get that,
$$
\E_{p(v_i^0,v_i^+)} f(v_i^0)^Tf(v_i^+)\leq \E_{p(v_i^0,v_i^1,v_i^2)} f(v_i^1)^Tf(v_i^2)+3\delta_{aug}^2.
$$

(c) $f(v_i^0)^Tf(v_i^1)\leq 1$, $f(v_i^0)^Tf(v_i^2)\leq 1$, and $\E_{p(v_i^1,v_i^2)} f(v_i^1)^Tf(v_i^2)\geq \frac{2-\E_{p(v_i^1,v_i^2)} ||f(v_i^1)-f(v_i^2)||^2}{2} \geq 1-\frac{\E_{p(v_i^1,v_i^2)} (||f(v_i^1)-f(v_i^0)||+||f(v_i^0)-f(v_i^2)||)^2}{2}\geq 1-2\delta_{aug}^2$.

\begin{lemma}[\citep{reversed_jensen} Corollary 3.5 (restated)] \label{lemma: reversed_jensen}

    Let $g:\mathbb{R}^m \rightarrow \mathbb{R}$ be a differentiable convex mapping and $z \in \mathbb{R}^n$. Suppose that $g$ is $L$- smooth with the constant $L > 0$, i.e. $\forall x,y\in\mathcal{R}^m, \Vert \nabla g(x) - \nabla g(y) \Vert \leq L \Vert x-y \Vert$. Then we have
    \begin{align*}
        0 \leq \E_{p(z)}g(z) - g\left(\E_{p(z)}z\right)\leq L\left[ \E_{p(z)} \lVert z\rVert^2 - \Vert \E_{p(z)} z\Vert^2\right]= L \sum_{j=1}^{n}\var(z^{(j)}),
    \end{align*}
    where $z^{(j)}$ denotes the $j$-th dimension of $v$.
\end{lemma}
\begin{lemma}[\citep{chaos} Lemma A.2. restated]
\label{lemma:Monte Carlo estimation}
    For ${\rm LSE}:=\log\E_{p(z)}\exp (f(v)^\top g(z))$, we denote its (biased) Monte Carlo estimate with $M$ random samples $z_i\sim p(z),i=1,\dots,M$ as $\widehat{\rm LSE}_M=\log\frac{1}{M}\sum_{i=1}^M \exp (f(v)^\top g(z_i))$. Then the approvimation error $A(M)$ can be upper bounded in expectation as 
    \begin{equation}
        A(M):=\E_{p(v,z_i)}|\widehat{\rm LSE}(M) - {\rm LSE}|\leq\mathcal{O}(M^{-1/2}).
        \nonumber
    \end{equation}
    We can see that the approvimation error converges to zero in the order of  $M^{-1/2}$. 
\end{lemma}
\begin{align*}
    \Lambda_2
    &=\E_{p(v_i^0)}\log\sum_{j=1}^K\exp(f(v_i^0)^T\mu_{y_j}) \\
    &=\E_{p(v_i^0)}\log\frac{1}{K}\sum_{i=j}^K\exp(f(v_i^0)^T\mu_{y_j})+\log K \\
    &=\E_{p(v_i^0)}\log\E_{p(y_j)}\exp(f(v_i^0)^T\mu_{y_j})+\log K \\
    &\overset{(d)}\geq \E_{p(v_i^1)}\log\E_{p(y_j)}\exp(f(v_i^1)^T\mu_{y_j})-\delta_{aug}-e\sum_{j=1}^n\var(\mu_j)+\log K \\
    &\overset{(e)}\geq \E_{p(v_i^1)}\E_{p(y_i)}\log\frac{1}{M}\sum_{j=1}^M\exp(f(v_i^1)^T\mu_{y_j})-A(M)+\log K-\delta_{aug}-e\sum_{j=1}^n\var(\mu_j) \\
    &=\E_{p(v_i^1)}\E_{p(y_i)}\log\frac{1}{M}\sum_{j=1}^M\exp(\E_{p(v_i^-|y_i^-)}f(v_i^1)^Tf(v_i^-))-A(M)+\log K-\delta_{aug}-e\sum_{j=1}^n\var(\mu_j) \\
    &\overset{(f)}\geq \E_{p(v_i^1)}\E_{p(y_i)}\E_{p(v_i^-|y_i^-)}\log\frac{1}{M}\sum_{i=1}^M\exp(f(v_i^1)^Tf(v_i^-)) \\
    &\hspace{1.5em}-\frac{1}{2}\sum_{j=1}^m\var(f_j(v^-|y))-A(M)+\log K-\delta_{aug}-e\sum_{j=1}^n\var(\mu_j)\\
    &=\E_{p(v_i^1)}\E_{p(y_i)}\E_{p(v_i^-|y_i^-)}\log\sum_{i=1}^M\exp(f(v_i^1)^Tf(v_i^-))\\
    &\hspace{1.5em}-\log M-\frac{1}{2}\sum_{j=1}^m\var(f_j(v^-|y))-A(M)+\log K-\delta_{aug}-e\sum_{j=1}^n\var(\mu_j).
\end{align*}

(d) We can show that:
$\exp(\left[f(v)^T\mu_{y_j}\right]$ is convex, and $u_{y_j}$ satisfy e-smooth,
\begin{align*}
    &\hspace{1.5em} ||\frac{\partial \exp(f(v)^Ta)}{\partial a}-\frac{\partial \exp(f(v)^Tb)}{\partial b}|| \\
    &=||\exp(f(v)^Ta)f(v)-\exp(f(v)^Tb)f(v))|| \\
    &=|\exp(f(v)^Ta)-\exp(f(v)^Tb)|\cdot||f(v)|| \\
    &\leq |\exp(f(v)^Ta)-\exp(f(v)^Tb)| \\
    &\leq e||(f(v)^T)(a-b)|| \quad (f(v)^Ta,f(v)^Tb \leq 1, \text{so the biggest slope is} \ e) \\
    &\leq e||a-b||. \\
\end{align*}
So according to Lemma \ref{lemma: reversed_jensen}, we get,
\begin{equation}
    \begin{split}
        \E_{p(y_j)}\exp(\left[f(v_i^1)^T\mu_{y_j}\right])
        &\leq \exp(\left[f(v_i^1)^T\E_{p(y_j)}\mu_{y_j}\right])+e\sum_{j=1}^n\var(\mu_j) \\
        &=\exp(f(v_i^1)^T\mu)+e\sum_{j=1}^n\var(\mu_j).
    \end{split}
    \nonumber
\end{equation}
Then, we can calculate the difference between $\log\E_{p(y_j)}\exp(\left[f(v_i^0)^T\mu_{y_j}\right])$ and $\log\E_{p(y_j)}\exp(\left[f(v_i^1)^T\mu_{y_j}\right])$ by applying reversed Jensen and Jensen inequality, respectively.
\begin{align*}
    &\hspace{1.5em} \log\E_{p(y_j)}\exp(\left[f(v_i^1)^T\mu_{y_j}\right])-\log\E_{p(y_j)}\exp(\left[f(v_i^0)^T\mu_{y_j}\right]) \\
    &\leq \log\E_{p(y_j)}\exp(\left[f(v_i^1)^T\mu_{y_j}\right])-\left[f(v_i^0)^T\mu\right] \\
    &\leq \log\left[\exp(f(v_i^1)^T\mu)+e\sum_{j=1}^n\var(\mu_j)\right]-\left[f(v_i^0)^T\mu\right]\\
    &=\log\left[\exp(f(v_i^1)^T\mu)\right]+\log\left[1+\frac{e\sum_{j=1}^n\var(\mu_j)}{\exp(f(v_i^1)^T\mu)}\right]-\left[f(v_i^0)^T\mu\right] \\
    &\leq f(v_i^1)^T\mu-f(v_i^0)^T\mu+\log\left[1+e\sum_{j=1}^n\var(\mu_j)\right] \quad (e^2\sum_{j=1}^n\var(\mu_j)\text{, if not ReLU)}\\
    &\leq (f(v_i^1)^T-f(v_i^0)^T)\mu+e\sum_{j=1}^n\var(\mu_j) \\
    &\leq (f(v_i^1)-f(v_i^0))^T\frac{||\mu||}{||f(v_i^1)-f(v_i^0)||}(f(v_i^1)-f(v_i^0))+e\sum_{j=1}^n\var(\mu_j) \\
    &\leq (f(v_i^1)-f(v_i^0))^T\frac{1}{||f(v_i^1)-f(v_i^0)||}(f(v_i^1)-f(v_i^0))+e\sum_{j=1}^n\var(\mu_j) \\
    &\leq \delta_{aug}+e\sum_{j=1}^n\var(\mu_j).
\end{align*}

(e) We adopt a Monte Carlo estimation with $M$ samples from $p(y)$ and bound the approximation error with Lemma \ref{lemma:Monte Carlo estimation}.

(f) We also uses Lemma \ref{lemma: reversed_jensen}, and as proof by \citet{chaos}, logsumexp is L-smooth for $L=\frac{1}{2}$.

\begin{align*}
        \gL_{\mathrm{CE}}
        &=\Lambda_1+\Lambda_2 \\
        &\geq-\E_{p(v,y)}f(v_i^1)^Tf(v_i^2)-3\delta_{aug}^2-\delta_{aug}-\E_{p(v^0,y)}||f(v_y^0)-\mu_y|| \\
        &\hspace{1.5em}+\E_{p(v_i^1)}\E_{p(y_i)}\E_{p(v_i^-|y_i)}\log\sum_{i=1}^M\exp(f(v_i^1)^Tf(v_i^-)) \\
        &\hspace{1.5em}-\log M-\frac{1}{2}\sum_{j=1}^m\var(f_j(v^-|y))-A(M)+\log K-\delta_{aug}-e\sum_{j=1}^n\var(\mu_j) \\
        &=\left[-\E_{p(v_i^1,v_i^2)}f(v_i^1)^Tf(v_i^2)+\E_{p(v_i^-)}\log\sum_{i=1}^M\exp(f(v_i^1)f(v_i^-))\right]-3\delta_{aug}^2-\delta_{aug}-\E_{p(v^0,y)}||f(v_y^0)-\mu_y||\\
        &\hspace{1.5em}-\log M-\frac{1}{2}\sum_{j=1}^m\var(f_j(v^-|y))-A(M)+\log K-\delta_{aug}-e\sum_{j=1}^n\var(\mu_j) \\
        &=\gL_{\mathrm{NCE}}-3\delta_{aug}^2-2\delta_{aug}-\log\frac{M}{K}-\frac{1}{2}\sum_{j=1}^m\var(f_j(v^-|y))-A(M)-e\sum_{j=1}^n\var(\mu_j)-\E_{p(v^0,y)}||f(v_y^0)-\mu_y|| \\
        &\overset{(g)}{\geq} \gL_{\mathrm{NCE}}-3\delta_{aug}^2-2\delta_{aug}-\log\frac{M}{K}-\frac{1}{2}\var(f(v^+)|y)-\sqrt{\var(f(v^0)|y)}-O(M^{-\frac{1}{2}})-e\var(\mu_y). \\
    \nonumber
\end{align*}
(g) This holds because, $v^-$ is randomly selected from $v^+$ and,
\begin{align*}
    &\hspace{1.5em}\sum_{j=1}^{m}\var(f_j(v^-|y))\\
    &=\sum_{j=1}^m\E_{p(y)}\E_{p(v|y)}(f_j(v^+)-\E_{p(v'|y)}f_j(v'))^2\\
    &=\E_{p(y)}\E_{p(v|y)}\sum_{j=1}^m(f_j(v^+)-\E_{v'}f_j(v'))\\
    &=\E_{p(y)}\E_{p(v|y)}||f(v)-\E_{v'}f(v')||^2\\
    &=\var(f(v^+)|y).
\end{align*}
And similarly, we can get $\sum_{j=1}^n\var(\mu_j)=\var(\mu_y)$. So the lower bound is proved.

\subsection{Proof of Corollary \ref{corollary:CE_MI}}\label{proof:corollary CE_MI}
For $\var(f(v_y^0|y))$, we can use augmentation distance and the intrinsic property of model and data to express.
\begin{align*}
    \var(f(v_y^0|y))
    &=\E_{p(y)}\E_{p(v_y^0|y)}||f(v_y^0)-\mu_y||^2\\
    &=\E_{p(y)}\E_{p(v_y^0|y)}\left[(f(v_y^0)-\mu_y)^T(f(v_y^0)-\mu_y))\right]\\
    &=\E_{p(y)}\E_{p(v_y^0|y)}\left[f(v_y^0)^Tf(v_y^0)+\mu_y^T\mu_y-2f(v_y^0)^T\mu_y\right]\\
    &\leq \E_{p(y)}\E_{p(v_y^0|y)}\left[2-2f(v_y^0)^T\mu_y\right]\\
    &\overset{(h)}{\leq} \E_{p(y)}\E_{p(v_y^0|y)}\left[2-2(1-\frac{1}{3}\delta_{aug}^2-\frac{2}{3}\delta_{aug}\delta_{y^+}-\frac{1}{2}\delta_{y^+}^2)\right]\\
    &=\E_{p(y)}\E_{p(v_y^0|y)}\left[\frac{2}{3}\delta_{aug}^2+\frac{4}{3}\delta_{aug}\delta_{y^+}+\delta_{y^+}^2)\right]\\
    &\leq \frac{2}{3}\delta_{aug}^2+\frac{4}{3}\delta_{aug}L\epsilon_0+L^2\epsilon_0^2,
\end{align*}
where $\epsilon_0=\E_{p(y)}\E_{p(v_i^0,v_j^0|y)}||v_i^0-v_j^0||$ and $L$ is the Lipschitz constant, so $\delta_{y^+}^2=\E_{p(y,i,j)}||f(v^0_{y,i})-f(v^0_{y,j})||^2 \leq (L\epsilon_0)^2$.

Then we can easily get that,
\begin{equation}
    \begin{split}
        \var(f(v_y^+)|y)
        &=\E_{p(y)}\E_{p(v_y^-|y)}||f(v_y^+)-\mu_y||^2\\
        &\leq \E_{p(y)}\E_{p(v_y^+|y)}(||f(v_y^+)-f(v_y^0)||+||f(v_y^0)-\mu_y||)^2\\
        &=\E_{p(y)}\E_{p(v_y^+|y)}||f(v_y^+)-f(v_y^0)||^2+\E_{p(y)}\E_{p(v_y^+|y)}||f(v_y^0)-\mu_y||)^2\\
        &\hspace{1.5em}+2\E_{p(y)}\E_{p(v_y^+|y)}||f(v_y^+)-f(v_y^0)||\cdot||f(v_y^0)-\mu_y||\\
        &\leq \delta_{aug}^2+\var(f(v_y^0)|y)+2\delta_{aug}\sqrt{\var(f(v_y^0)|y)}\\
        &= (\delta_{aug}+\sqrt{\var(f(v_y^0)|y)})^2.
    \end{split}
    \nonumber
\end{equation}

(h) We use Theorem \ref{theorem:aug_cla}.

And $\var(\mu_y)$ can also be expressed by intrinsic properties.
\begin{align*}
    \var(\mu_y)
    &=\E_{p(y)}||\mu_y-\mu||^2\\
    &=\E_{p(y)}||\mu_y-f(v_y^{*})+f(v_y^{*})-\mu||^2 \\
    &\leq \E_{p(y)}(||\mu_y-f(v_y^{*})||+||f(v_y^{*})-\mu||)^2 \\
    &=\E_{p(y)}||\E_{p(v_y|y)}f(v_y)-f(v_y^{*})||^2+\E_{p(y)}||f(v_y^{*})-\E_{p(v)}f(v)||^2\\
    &\hspace{1.5em}+2\E_{p(y)}(||\E_{p(v_y|y)}f(v_y)-f(v_y^{*})||\cdot||f(v_y^{*})-\E_{p(v)}f(v)||) \\
    &=\E_{p(y)}||\E_{p(v_y|y)}[f(v_y)-f(v_y^{*})]||^2+\E_{p(y)}||\E_{p(v)}[f(v_y^{*})-f(v)]||^2\\
    &\hspace{1.5em}+2\E_{p(y)}(||\E_{p(v_y|y)}[f(v_y)-f(v_y^{*})]||\cdot||\E_{p(v)}[f(v_y^{*})-f(v)]||) \\
    &\leq \E_{p(y)}\E_{p(v_y|y)}||f(v_y)-f(v_y^{*})||^2+\E_{p(y)}\E_{p(v)}||f(v_y^{*})-f(v)||^2\\
    &\hspace{1.5em}+2\E_{p(y)}(\E_{p(v_y|y)}||f(v_y)-f(v_y^{*})||\cdot||f(v_y^{*})-f(v)||)\\
    &\leq L^2\epsilon_1^2+L^2\epsilon_2^2+2L^2\epsilon_1\epsilon_2\\
    &=L^2(\epsilon_1+\epsilon_2)^2,
\end{align*}
where $v_y^*$ could be any node of class $y$, and $\epsilon_1=\E_{p(v,y)}||v_y-v_y^*||$, $\epsilon_2=\E_{p(y)}\E_{p(v)}||v-v_y^*||$.

\begin{align*}
    \hat{\gL}_{\mathrm{CE}}
    &\geq \gL_{\mathrm{NCE}}-3\delta_{aug}^2-2\delta_{aug}-\log\frac{M}{K}-\frac{1}{2}\var(f(v^-)|y)-\sqrt{\var(f(v^0)|y)}-O(M^{-\frac{1}{2}})-e\var(\mu_y)\\
    &\geq \gL_{\mathrm{NCE}}-3\delta_{aug}^2-2\delta_{aug}-\log\frac{M}{K}-\frac{1}{2}(\delta_{aug}+\sqrt{\var(f(v_y^0)|y)})^2-\sqrt{\var(f(v_y^0)|y)}-O(M^{-\frac{1}{2}})-eL^2(\epsilon_1+\epsilon_2)^2\\
    &=\gL_{\mathrm{NCE}}-3\delta_{aug}^2-2\delta_{aug}-log\frac{M}{K}-\frac{1}{2}\delta_{aug}^2-(\delta_{aug}+1)\sqrt{\var(f(v_y^0)|y))}\\
    &\hspace{1.5em}-\frac{1}{2}\var(f(v_y^0)|y))-O(M^{-\frac{1}{2}})-eL^2(\epsilon_1+\epsilon_2)^2\\
    &=\gL_{\mathrm{NCE}}-g(\delta_{aug})-\log\frac{M}{K}-O(M^{-\frac{1}{2}}),
\end{align*}
where $g(\delta_{aug})=\frac{23}{6}\delta_{aug}^2+\frac{1}{2}L^2\epsilon_0^2+eL^2(\epsilon_1+\epsilon_2)^2+2\delta_{aug}+\frac{2}{3}\delta_{aug}L\epsilon_0+(\delta_{aug}+1)\sqrt{\frac{2}{3}\delta_{aug}^2+\frac{4}{3}\delta_{aug}L\epsilon_0+L^2\epsilon_0^2}$.

According to \citet{NCE_MI}, we get,
\begin{equation}
    \begin{split}
        &I(f(v_i^1);f(v_i^2)) \geq \log(M)-\gL_{\mathrm{NCE}}, \\
        &\gL_{\mathrm{NCE}} \geq \log(M)-I(f(v_i^1);f(v_i^2)).
    \end{split}
    \nonumber
\end{equation}
Therefore, we can reformulate Theorem \ref{theorem:generalization} as below:
\begin{equation}
    \begin{split}
        \hat{\gL}_{\mathrm{CE}}
        &\geq \log(M)-I(f(v_i^1);f(v_i^2))-g(\delta_{aug})-\log\frac{M}{K}-O(M^{-\frac{1}{2}})\\
        &=\log(K)-I(f(v_i^1);f(v_i^2))-g(\delta_{aug})-O(M^{-\frac{1}{2}}).
    \end{split}
    \nonumber
\end{equation}

\subsection{Proof of Corollary \ref{corollary:spectrum_positive_pair_difference}} \label{proof:corollary_spectrum_positive_pair_difference}
    Corallary \ref{corollary:spectrum_positive_pair_difference} could  be simply proved below:
    \begin{equation}
        \begin{split}
            \E_{p(v_i^1,v_i^2)}||f(v_i^1)-f(v_i^2)||^2
            &=\E_{p(v_i^1,v_i^2)}[(f(v_i^1)^T-f(v_i^2)^T)(f(v_i^1)-f(v_i^2))]\\
            &=\E_{p(v_i^1,v_i^2)}[2-2f(v_i^1)^Tf(v_i^2)] \\
            &=2-\frac{2} {N}tr((H^1)^TH^2)\\
            &\overset{(1)}{=}2-\frac{2}{N}\sum_i\theta_i\lambda_i'\lambda_i''. \\
        \end{split}
        \nonumber
    \end{equation}
    So $(\E_{p(v_i^1,v_i^2)}||f(v_i^1)-f(v_i^2)||)^2\leq \E_{p(v_i^1,v_i^2)}||f(v_i^1)-f(v_i^2)||^2=2-\frac{2}{N}\sum_i\theta_i\lambda_i'\lambda_i''$, then $\E_{p(v_i^1,v_i^2)}||f(v_i^1)-f(v_i^2)|| \leq \sqrt{2-\frac{2}{N}\sum_i\theta_i\lambda_i'\lambda_i''}$.
    
    (1) is suggested by \citet{GCL_specturm_NCE}, $tr((H^1)^TH^2)$ could be represented as $\sum_i\theta_i\lambda_i'\lambda_i''$.

    As we know that,
    \begin{equation}
        \begin{split}
            \E_{p(v_i^1,v_i^2)}||f(v_i^1)-f(v_i^2)||\leq \E_{p(v_i^1,v_i^2)}(||f(v_i^1)-f(v_i^0)||+||f(v_i^0)-f(v_i^2)||)\leq 2\delta_{aug}.
        \end{split}
        \nonumber
    \end{equation}
    Then, we can get:
    \begin{equation}
        2\delta_{aug}\geq \E_{p(v_i^1,v_i^2)}||f(v_i^1)-f(v_i^2)|| \geq \sqrt{2-\frac{2}{N}\sum_i\theta_i\lambda_i'\lambda_i''}.
    \end{equation}

\subsection{Proof of Lemma \ref{lemma:change_of_spectrum}}\label{proof:lemma change of spectrum}
From \citet{mtxper}, we know the following equation:
\begin{equation}
\Delta \lambda_i=\lambda_i^{'}-\lambda_i=u_i^T\Delta A u_i-\lambda_i u_i^T\Delta D u_i+O(||\Delta A||).
\nonumber
\end{equation}
So we can calculate the difference between $\lambda_i',\lambda_i''$ and $\lambda_i$,
\begin{align*}
    \Delta \lambda_i
    &=\sum_m (\sum_n u_i\left[n\right]\Delta A\left[m\right]\left[n\right])u_i\left[m\right]-\lambda_i \sum_m u_i\left[m\right]\Delta D\left[m\right] u_i\left[m\right]+O(||\Delta A||)\\
    &=\sum_{m,n}u_i\left[m\right]u_i\left[n\right]\Delta A\left[m\right]\left[n\right]-\lambda_i \sum_{m,n}u_i\left[m\right]\Delta A\left[m\right]\left[n\right]u_i\left[m\right]+O(||\Delta A||).
\end{align*}

And we can directly calculate $\lambda_i'-\lambda_i''$ as below:
\begin{align*}
        \lambda_i'-\lambda_i''
        &=\Delta \lambda_i'-\Delta \lambda_i'' \\
        &=\sum_{m,n}u_i\left[m\right]u_i\left[n\right]\Delta \widehat{A}\left[m\right]\left[n\right]-\lambda_i \sum_{m,n}u_i\left[m\right]\Delta \widehat{A}\left[m\right]\left[n\right]u_i\left[m\right]\\
        &=\sum_{m,n}u_i\left[m\right]\Delta\widehat{A}\left[m\right]\left[n\right](u_i\left[n\right]-\lambda_i u_i\left[m\right]).
\end{align*}

\section{GCL Methods with Spatial and Spectral Augmentation}
\subsection{Spatial Augmentation} \label{appendix:spatial augmentation and MI}
    Most augmentation methods are applied to explicitly or implicitly increase mutual information while maintain high augmentation distance. GRACE simply adjusts this by changing the drop rate of features and edges. AD-GCL \citep{GCL_AD_GCL} directly uses the optimization objective $min_{\{aug\}}max_{\{f \in F\}}I(f(v);f(aug(v)))$ to search for a stronger augmentation.
    \begin{wrapfigure}{r}{7cm}
        \centering
        \includegraphics[width=0.4\textwidth]{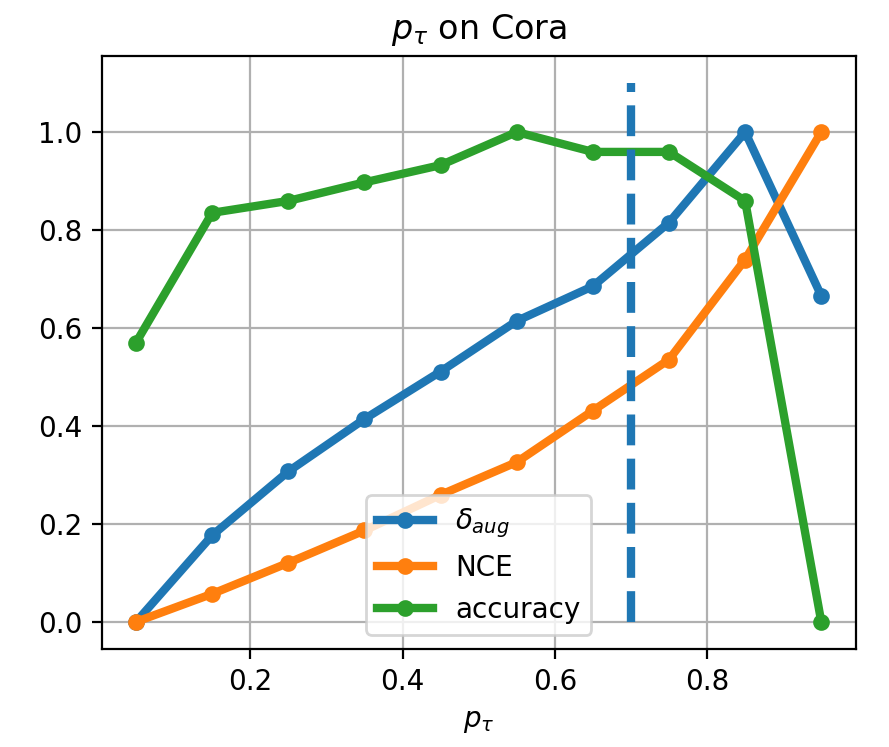}
        \caption{influence of $p_{\tau}$ on Cora (all the data are normalized for better visualization)}
        \label{fig:GCA threshold on cora}
    \end{wrapfigure}
    
    And GCA \citep{GCL_GCA} could always perform better than random drop. This is mainly because GCA calculates node importance and masks those unimportant to increase mutual information. Also they use $p_{\tau}$ as a cut-off probability, so for those unimportant features/edges, all of them share the same drop probability $p_{\tau}$. By setting a large $p_{\tau}$, GCA can reduce the drop probability for the least important features/edges and drop more relatively important ones to achieve a trade-off between mutual information and augmentation distance.

    From Figure \ref{fig:GCA threshold on cora}, we could clearly see that, as $p_{\tau}$ increases, augmentation distance and $\gL_{\mathrm{NCE}}$ are increasing, and leads to a better downstream performance, than when $p_{\tau}$ becomes too large, we got a trivial solution. And in the code provided by the author, $p_{\tau}$ is set to 0.7. So GCA performances well on downstream tasks not only because its adaptive augmentation, but also its modification on augmentation distance.
    
\subsection{Spectral Augmentation}\label{appendix:spectral augmentation and lambda}

    Furthermore, we can demonstrate that lots of spectral augmentations follow this schema to improve downstream performance. \citet{GCL_specturm_NCE} proposes that increasing the number of high-frequency drops leads to better performance. This is because high-frequency parts are associated with higher coefficients $\lambda_i$, so increasing the number of high-frequency drops can have a stronger increasement on $\delta_{aug}$, resulting in better performance.

    \begin{lemma}[Change of Spectrum] \label{lemma:change_of_spectrum}
    if we assume that $A'=A+\Delta A_1$, $A''=A+\Delta A_2$, $\lambda_i'$, $\lambda_i''$ is the $i^{th}$ eigenvalue of $A'$ and $A''$, respectively. $\Delta \hat{A} = A'- A''$, and $u_i$ is the corresponding eigenvector. 
        \begin{equation}
            \begin{split}
                \lambda_i'-\lambda_i''=\sum_{m,n}u_i\left[m\right]\Delta\widehat{A}\left[m\right]\left[n\right](u_i\left[n\right]-\lambda_i u_i\left[m\right]).
            \end{split}
            \nonumber
        \end{equation}
    \end{lemma}

    Lemma \ref{lemma:change_of_spectrum} is proved in Appendix \ref{proof:lemma change of spectrum}. \citet{GCL_max_spectrum} propose to maximize the spectral difference between two views, but Lemma \ref{lemma:change_of_spectrum} shows that difference between spectrum is highly correlated with the original magnitude, so it is actually encouraging more difference in large $|\lambda_i|$. But rather than just drop information, they try to improve the spectrum of first view, and decrease the other view. if we simply assume $\lambda_i'=\lambda_i+n$, $\lambda_i''=\lambda_i-n$, then $\lambda_i'\lambda_i''=\lambda_i^2-n^2\leq \lambda_i^2$, so this could also be explained by augmentation distance increasement.

\section{Further Explanation} \label{further_explanation}
\subsection{Value of $\theta$s}\label{appendix:value of thetas}

    As defined by \citet{GCL_specturm_NCE}, $\theta$s are actually linear combination of the eigenvalues of adjacency matrix $\mA$. To demonstrate what $\theta$s actually are, we first focus on the assumption below.
    \begin{assumption}[High-order Proximity]
        $\bm{M}=w_0+w_1\bm{A}+w_2\bm{A^2}+\dots+w_q\bm{A^q}$, where $\bm{M}=X^1W\cdot W^T(X^2)^T$, $\bm{A^i}$ means matrix multiplications between $i$ $\bm{A}$s, and $w_i$ is the weight of that term.
    \end{assumption}
    Where $X^1,X^2$ indicates the feature matrix of graph $\gG^1,\gG^2$, $W$ stands for the parameter of the model, so  $\bm{M}=X^1W\cdot W^T(X^2)^T$ means embedding similarity between two views, and could be roughly represented by the weighted sum of different orders of $\bm{A}$. Furthermore, we have that:

    \begin{equation}
        \label{set}
        \left\{
        \begin{aligned}
        w_0+w_1\lambda_1+&\dots+w_q\lambda_1^q=\theta_1\\
        w_0+w_1\lambda_2+&\dots+w_q\lambda_2^q=\theta_2\\
        &\dots\\
        w_0+w_1\lambda_N+&\dots+w_q\lambda_N^q=\theta_N,\\
        \end{aligned}
        \right.\\
        \nonumber
    \end{equation}
    where $\lambda_1,...,\lambda_N$ is N eigenvalues of the adjacency matrix $\mA$.

    \begin{figure}
        \centering
        \includegraphics[width=0.8\textwidth]{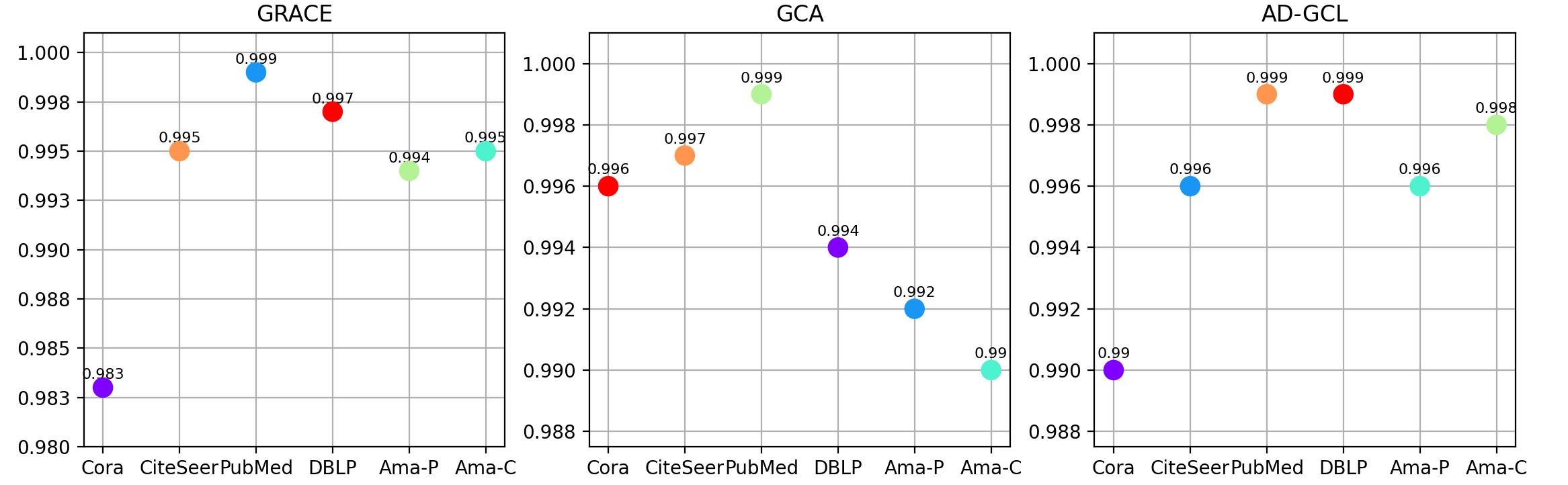}
        \caption{Percentage of positive $\theta$}
        \label{fig:percentage of positive}
    \end{figure}

    So we know that $\theta$s are actually linear combination of $\lambda$s. As the model is trained to minimize $\gL_{\mathrm{NCE}}$, $\theta$s are expected to increase, and we can simply set $w_0,w_2,...,w_{2(q/2)}$ to be positive and other $w_i$ to $0$, then we can get $\theta$s that are all positive, and the model would easily find better $w$s.
    
    We can say that in the training process, $\theta$s are mostly positive, and the experiments shown in Figure \ref{fig:change of theta} indicate it true.
    
    \begin{wrapfigure}{r}{9cm}
        \centering
        \includegraphics[width=0.89\linewidth]{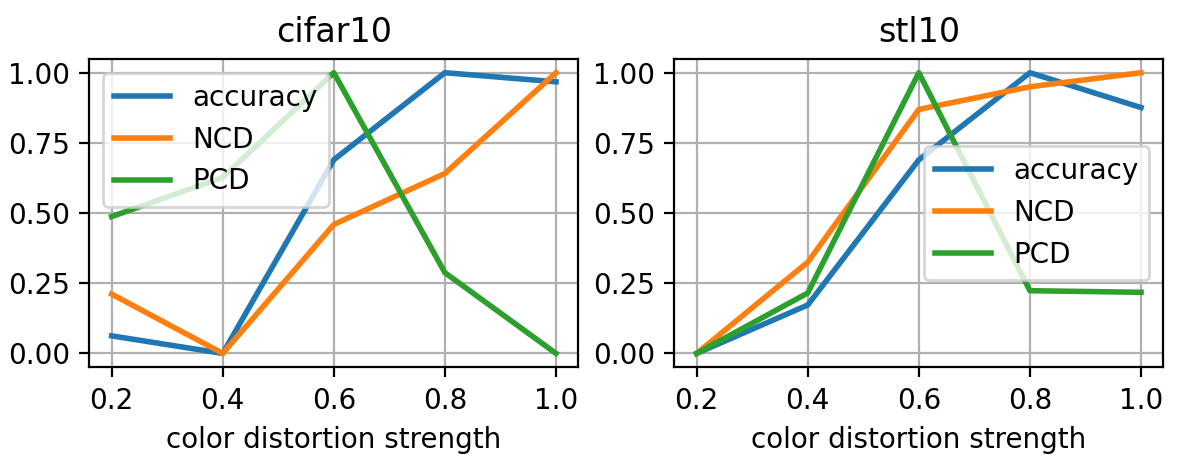}
        \caption{relationship of PCD, NCD and performance on images}
        \label{fig:CV_simi_simy}
    \end{wrapfigure}
\subsection{PCS, NCS and Downstream Performance} \label{appendix:more_PCS}

\begin{figure}
        \centering
        \includegraphics[width=0.9\textwidth]{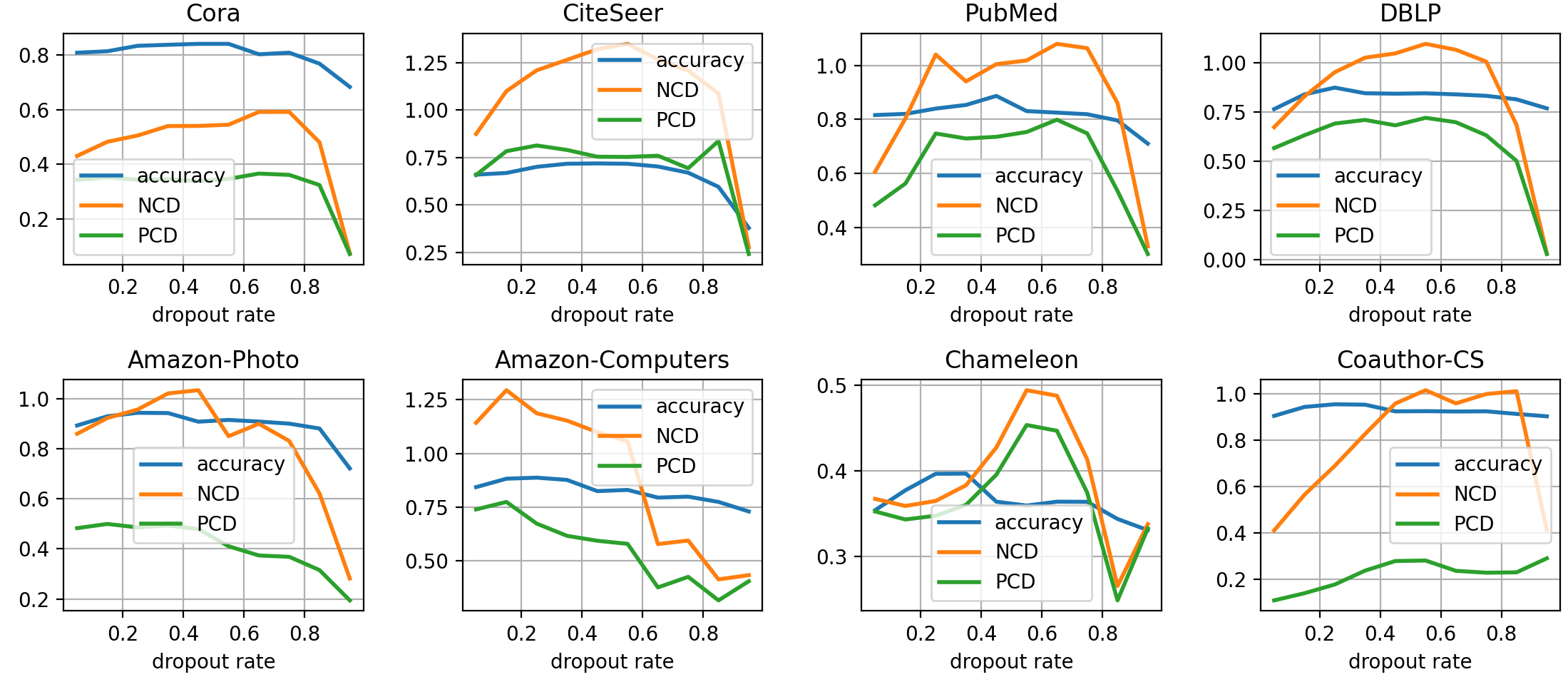}
        \caption{More experiments on PCS and NCS, the detailed data is slightly different due to randomness, but it shows similar tendency}
        \label{fig:simi_simy_all}
    \end{figure}

    More experiments are conducted on various of datasets to show that our finding could be generalized rather that limited to few datasets in Figure \ref{fig:simi_simy_all}. They show similar tendency that with the dropout rate increasing, the downstream accuracy increases first and decreases when the augmentation is too strong. And those experiments show that when the downstream accuracy increases, the positive center distance are sometimes increasing, and the better downstream performance is mainly caused by the increasing distance of negative center.

    We also conduct experiments on images to verify our theory, we control the magnitude of augmentation by adjusting the color distortion strength, and the results are normalized by Min-Max normalization. From Figure \ref{fig:CV_simi_simy}, we can observe that the downstream performance is also closely correlated with negative center distance especially when the color distortion strength changes from 0.2 to 0.6 the positive center distance increases while downstream performance is increasing, but when color distortion is greater than 0.6 the positive center distance also tends to decrease. This aligns with our finding in Theorem \ref{theorem:aug_cla} that with the augmentation gets stronger the negative center distance is increasing while the positive center distance does not change in specific pattern. Also the color distortion is not strong enough to change the label information, so the downstream performance keeps increasing with stronger augmentation.

\subsection{Change of $\delta_{aug}$ and Label Consistency} \label{appendix:delta_and_KL}

    \begin{wrapfigure}{r}{9cm}
        \centering
        \includegraphics[width=0.9\linewidth]{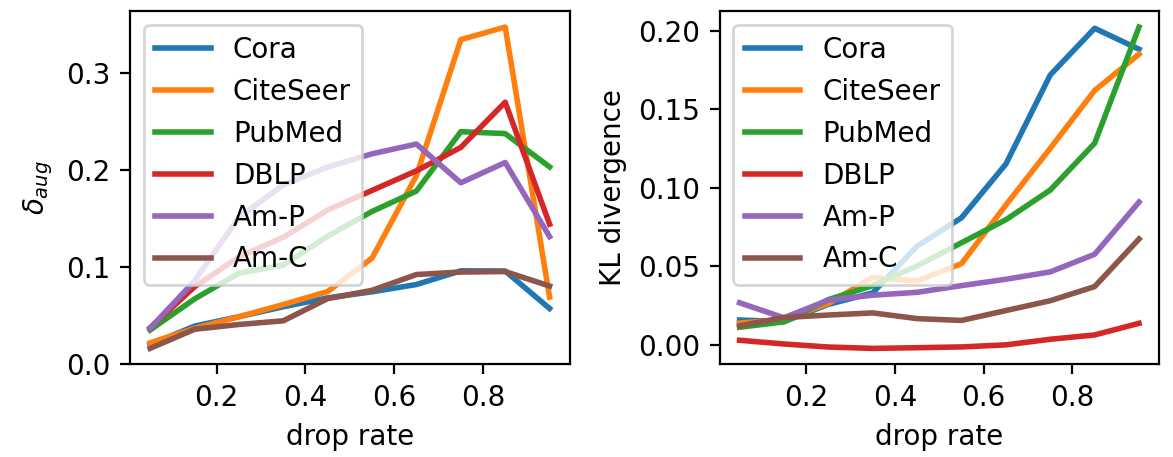}
        \caption{relationship between $\delta_{aug}$, KL divergence and augmentation}
        \label{fig:delta_aug and consistency}
    \end{wrapfigure}
    To verify how is $\delta_{aug}$ changing with stronger augmentation, we use drop rate of edges/features as data augmentation, and find that when the drop rate increases, $\delta_{aug}$ also tends to increase. Also to verify the view invariance assumption, we first train a well conditioned model and use its prediction as $p(v_i)$, then we change the drop rate and calculate new $p'(v_i)$, then we can observe from Figure \ref{fig:delta_aug and consistency} that though the KL divergence is increasing with drop rate, it remains quite small magnitude, so the label is consistent with data augmentation.

\subsection{Augmentation Free Methods} \label{appendix:augmentation_free_methods}

    In this paper, we mainly discuss how the augmentation will affect the contrastive performance, but actually, GCL methods with or without augmentation aim for the same, they both try to align intra-class nodes and separate inter-class nodes. However, during contrastive learning, label information is not accessible, so they use different methods to get intra-class nodes. 
    \begin{itemize}
        \item GCL methods with augmentation create intra-class nodes by data augmentation, so it is necessary to control the strength of augmentation to ensure label consistency. But augmentation brings more flexibility, you can freely change the topology and feature of the graph, so a good GCL method with augmentation always require a well-designed data augmentation. This could lead to great performance, but the they require more time consumption and overlook the unique properties of graphs.
        \item GCL methods without augmentation instead find intra-class nodes by other methods. For example, GMI \cite{GMI} and iGCL \cite{iGCL} try to align the anchor with its neighbors and similar nodes (which are more likely to hold the same label), and SUGRL \cite{SUGRL} create intra-class nodes by two different embedding generation methods. Label based methods like SupCon \cite{SupCon} directly align samples with the same class. These methods take advantage of the inherent property of the dataset such as homophily and the similarity between intra-class samples but the positive sample construction is not as flexible as augmentation.
    \end{itemize}
    Therefore, GCL methods with or without methods are inherently the same, they both align positive samples, and they create the positive samples differently. Our analysis focus on the difference between two positive samples, so the analysis can also be employed on those methods.
\subsection{change on positive/negative pair similarity}\label{appendix:change on positive/negative pair similarity}
\begin{figure}
        \centering
        \includegraphics[width=0.9\textwidth]{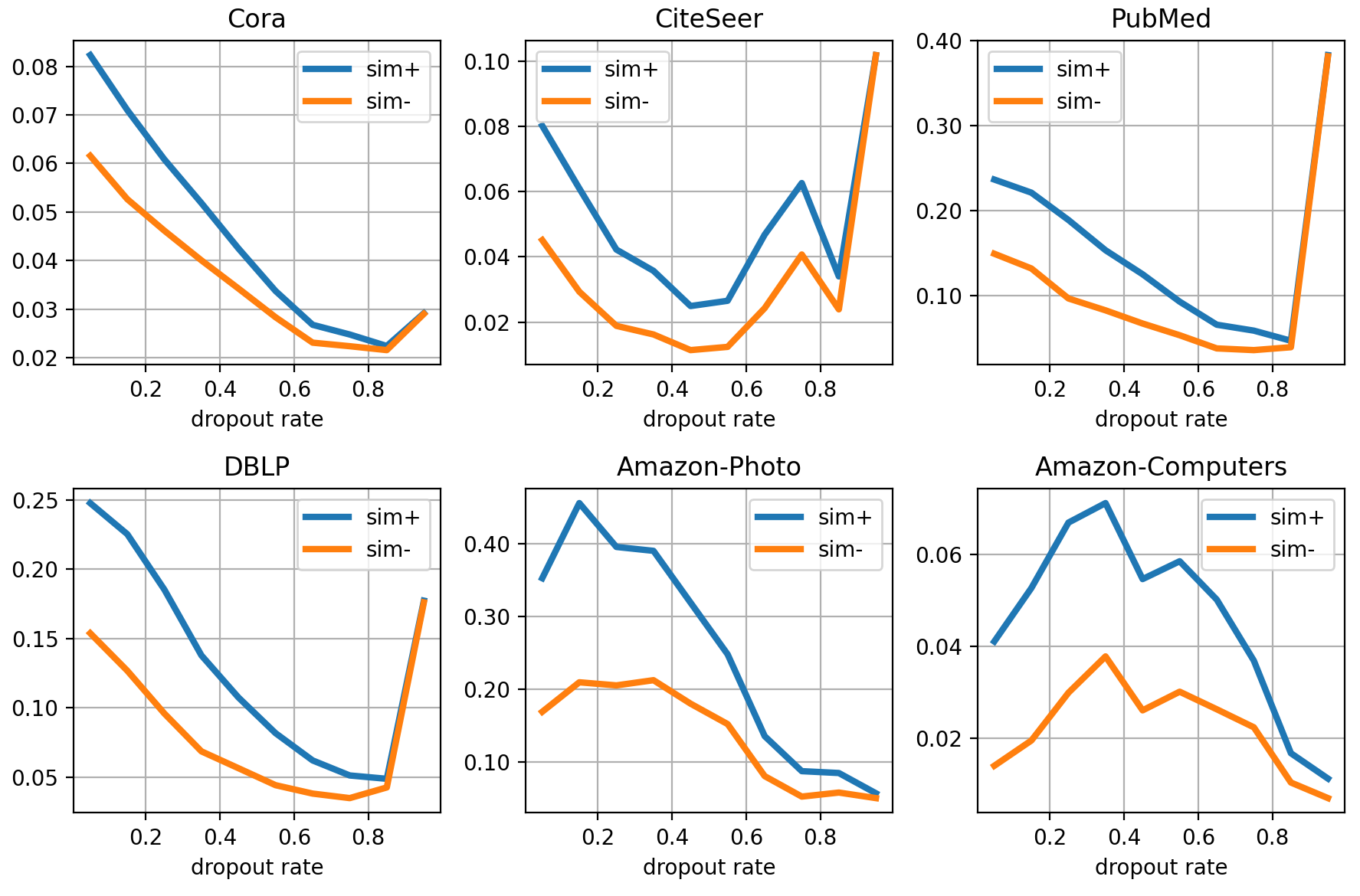}
        \caption{$\mathrm{sim+}$ represents the positive pair similarity $f(v_i^1)^Tf(v_i^2)$, and $\mathrm{sim-}$ is negative pair similarity $f(v_i^1)^Tf(v_i^-)$, the x-axis stands for dropout rate on edges}
        \label{fig:pos_neg_sim}
    \end{figure}

    The InfoNCE loss $\gL_{\mathrm{NCE}}$ can be written as $\gL_{\mathrm{NCE}}=\E_{p(v_i^1,v_i^2)}\E_{p({v_i^-})}\left[-\log\frac{\exp(f(v_i^1)^Tf(v_i^2))}{\sum_{\{ v^-_i\}}\exp(f(v_i^1)^Tf(v^-_i))}\right]$, and when we perform stronger augmentation, $f(v_i^1)^Tf(v_i^2)$ would be hard to maximize, and the model will try to minimize $f(v_i^1)^Tf(v^-_i)$ harder. From Figure \ref{fig:pos_neg_sim}, when the augmentation gets stronger, negative and positive pair similarity both decreases, so the class separating performance is enhanced.
    
\section{Experiments} \label{experiments}

\subsection{Datasets and Experimental Details} \label{appendix:datasets_baselines}
    We choose the six commonly used Cora, CiteSeer, PubMed, DBLP, Amazon-Photo and Amazon-Computer for evaluation. The first four datasets are citation networks \citep{dataset_Cora_CiteSeer_PubMed1,dataset_Cora_CiteSeer_PubMed2,dataset_DBLP}, where nodes represent papers, edges are the citation relationship between papers, node features are comprised of bag-of-words vector of the papers and labels represent the fields of papers. In Amazon-Photos and Amazon-Computers \citep{dataset_Amazon_P_C}, nodes represent the products and edges means that the two products are always bought together, the node features are also comprised of bag-of words vector of comments, labels represent the category of the product.

    We use 2 layers of GCNConv as the backbone of encoder, we use feature/edge drop as data augmentation, the augmentation is repeated randomly every epoch, and InfoNCE loss is conducted and optimized by Adam. After performing contrastive learning, we use logistic regression for downstream classification the solver is liblinear, and in all 6 datasets we randomly choose 10\% of nodes for training and the rest for testing. 
    \begin{table}[!htbp]
      \caption{Dataset statistics}
      \label{Dataset_statistics}
      \centering
      \setlength{\tabcolsep}{1mm}{
        \begin{tabular}{ccccc}
            \toprule 
            Dataset & Nodes & Edges & Features & Classes \\
            \midrule
            Cora  & 2,708 & 5,429 & 1,433 & 7 \\
            Citeseer  & 3,327 & 4,732 & 3,703 & 6\\
            Pubmed  & 19,717 & 44,338 & 500 & 3\\
            DBLP  & 17,716 & 105,734 &1,639 & 4\\
            Amazon-Photo  & 7,650 & 119,081 & 745 & 8 \\
            Amazon-Computers  & 13,752 & 245,861 & 767 & 10 \\
            \bottomrule
        \end{tabular}
    }
    \end{table}

    \begin{table*}[!htbp]
      \caption{Dataset download links}
      \centering
      \label{Dataset_download_links}
      \small
      \setlength{\tabcolsep}{3mm}{
            \begin{tabular}{cl}
            \toprule 
            Dataset & Download Link \\
            \midrule 
            Cora & \href{https://github.com/kimiyoung/planetoid/raw/master/data}{https://github.com/kimiyoung/planetoid/raw/master/data} \\
            Citeseer & \href{https://github.com//kimiyoung/planetoid/raw/master/data}{https://github.com//kimiyoung/planetoid/raw/master/data} \\
            Pubmed & \href{https://github.com/kimiyoung/planetoid/raw/master/data}{https://github.com/kimiyoung/planetoid/raw/master/data} \\
            DBLP & \href{https://github.com/abojchevski/graph2gauss/raw/master/data/dblp.npz}{https://github.com/abojchevski/graph2gauss/raw/master/data/dblp.npz} \\
            \midrule 
            Amazon-Photo & \href{https://github.com/shchur/gnn-benchmark/raw/master/data\\/npz/amazon\_electronics\_photo.npz}{https://github.com/shchur/gnn-benchmark/raw/master/data/npz/amazon\_electronics\_photo.npz}\\
            Amazon-Computers & \href{https://github.com/shchur/gnn-benchmark/raw/master/data/npz/amazon\_electronics\_computers.npz}{https://github.com/shchur/gnn-benchmark/raw/master/data/npz/amazon\_electronics\_computers.npz} \\
            \bottomrule
            \end{tabular}
        }
    \end{table*}
    And the publicly available implementations of Baselines can be found at the following URLs:
    \begin{itemize}
        \item GCN: \href{https://github.com/tkipf/gcn}{https://github.com/tkipf/gcn}
        \item GAT: \href{https://github.com/PetarV-/GAT}{https://github.com/PetarV-/GAT}
        \item GRACE: \href{https://github.com/CRIPAC-DIG/GRACE}{https://github.com/CRIPAC-DIG/GRACE}
        \item GCA: \href{https://github.com/CRIPAC-DIG/GCA}{https://github.com/CRIPAC-DIG/GCA}
        \item AD-GCL: \href{https://github.com/susheels/adgcl}{https://github.com/susheels/adgcl}
        \item GCS: \href{https://github.com/weicy15/GCS}{https://github.com/weicy15/GCS}
        \item SpCo: \href{https://github.com/liun-online/SpCo}{https://github.com/liun-online/SpCo}
    \end{itemize}

\subsection{Hyperparameter Setting}
    \begin{table*}[h]
      \caption{Hyperparameters settings}
      \centering
      \small
      \label{table:hyperparameter}
      \setlength{\tabcolsep}{1.8mm}{
        \begin{tabular}{cccccccccc}
        \toprule 
            Dataset & Learning rate & Weight decay & num layers &$\tau$ & Epochs & Hidden dim & Activation \\
            \midrule 
            Cora & $5^{-4}$ & $10^{-6}$ & 2 & 0.4 &200 &128 &ReLU\\
            Citeseer & $10^{-4}$ & $10^{-6}$ & 2 & 0.9 & 200 & 256 &PReLU\\
            Pubmed & $10^{-4}$ & $10^{-6}$ & 2 & 0.7 &200 &256 &ReLU\\
            DBLP & $10^{-4}$ & $10^{-6}$ &2 & 0.7 &200 &256 &ReLU\\
            \midrule 
            Amazon-Photo & $10^{-4}$ & $10^{-6}$ & 2 &0.3 &200 &256 &ReLU\\
            Amazon-Computers & $10^{-4}$ & $10^{-6}$ &2 &0.2 &200 &128 &RReLU\\
            \bottomrule
        \end{tabular}
        }
    \end{table*}

    The hyperparameter settings is shown in Table \ref{table:hyperparameter}, other hyperparameter correlated to only one algorithm are set the same as the original author. The hyperparameter in our methods retain rate $\xi$ and spectrum change magnitude $\alpha$, we select them from 0.05 to 0.45 and from -0.1 to 0.01, respectively.

\subsection{Changes on the Spectrum} \label{appendix:changes on the spectrum}

    \begin{figure}
        \centering
        \includegraphics[width=1\textwidth]{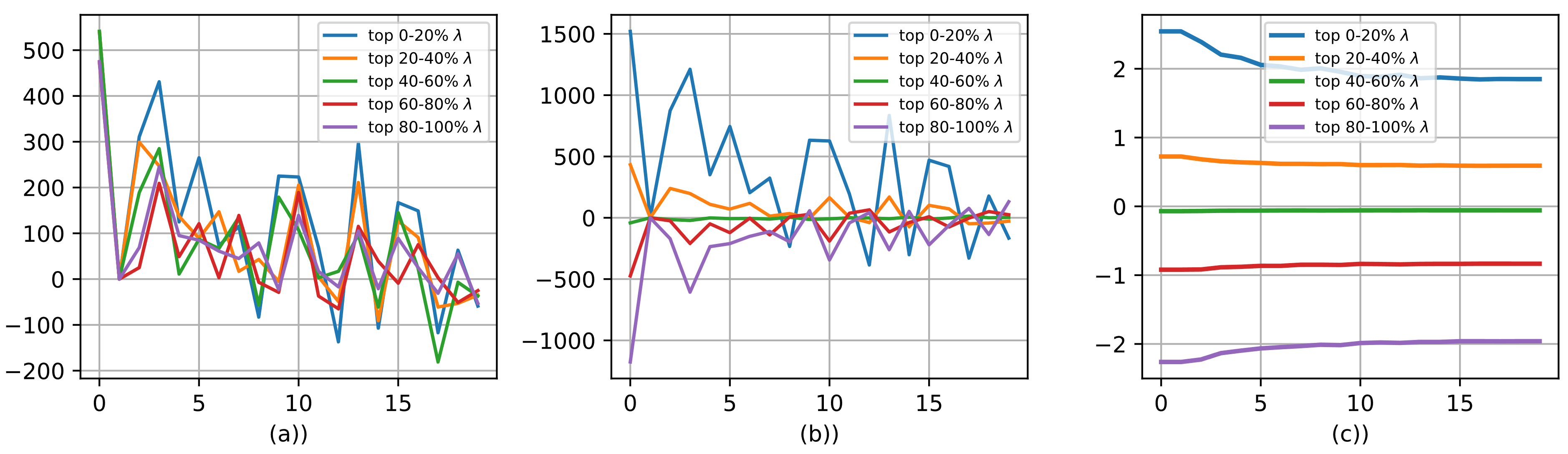}
        \caption{As we perform spectrum augmentation each 10 epochs, the x-axis is epoch/10, the y-axis of the left figure is number of decreasing $\lambda$s minus number of increasing $\lambda$s; for the middle one, y-axis stands for how much $\lambda$s averagely decreases; and the right one is the average value of $\lambda$.}
        \label{fig:change of theta}
    \end{figure}

    From Figure \ref{fig:change of theta}(a), we can see that, when the algorithm is training, $\theta$s are mostly increasing gradually, and when we perform spectrum augmentation, $\theta$s will not increase as before, increasing number of $\theta$ is close even smaller to decreasing ones. Then we take a step back on those decreasing ones, result in increasing $\theta$s again in the next epoch. Therefore, what we do is actually perform augmentation to maximize augmentation distance first, then maximize the mutual information after spectrum augmentation. The idea is actually similar AD-GCL, but we use $\theta$s to indicate whether the augmentation is too much, so we get a more reasonable result. Figure \ref{fig:change of theta}(b) and (c) shows that as the training goes, the change on larger magnitude eigenvalues are also more significant, causing the spectrum to be smoother.
    
    Also there is one thing to notice that when we perform spectrum smoothen method, we are indirectly changing the edge weights, causing the augmentation being weaker or stronger as drop an edge with weight of 1 is different than drop an edge with weight $1+noise$. To reduce its influence, we conduct extra augmentation or recovery based on the average weight change.

\subsection{Center Distance} \label{appendix:center_distance}
\begin{figure}[h]
        \centering
        \includegraphics[width=0.9\textwidth]{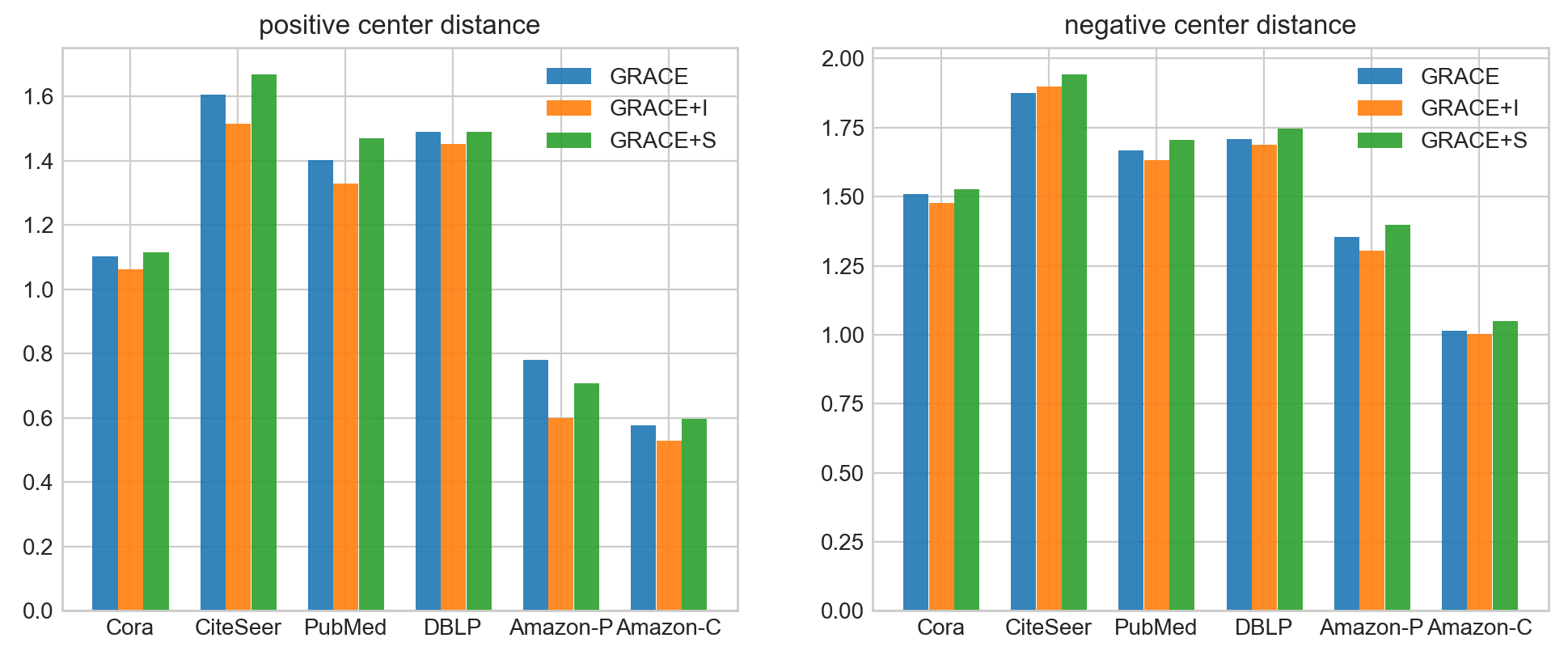}
        \caption{distance of nodes between its positive center and negative center, GRACE stands for the pure GRACE, GRACEI stands for GRACE with information augmentation, and GRACES stands for GRACE with spectrum augmentation}
        \label{fig:sim_GRACE}
    \end{figure}

    As we mentioned earlier, GCL mainly contributes to downstream tasks by increasing the negative center distance while maintaining a relatively small distance to the positive center. We propose two methods: one that increases mutual information between two views while keeping a high augmentation distance by masking more unimportant edges or features. This allows the model to learn more useful information, which forces nodes close to its positive center. The other method tries to increase augmentation distance while maintaining a relatively high mutual information, so it may not learn as much useful information. However, by increasing the augmentation distance, it forces the model to separate nodes from different classes further apart. In summary, the first method brings nodes of the same class closer together, while the second method separates nodes from different classes further apart just as shown in Figure \ref{fig:sim_GRACE}.

    \begin{figure}[h]
        \centering
        \includegraphics[width=0.9\textwidth]{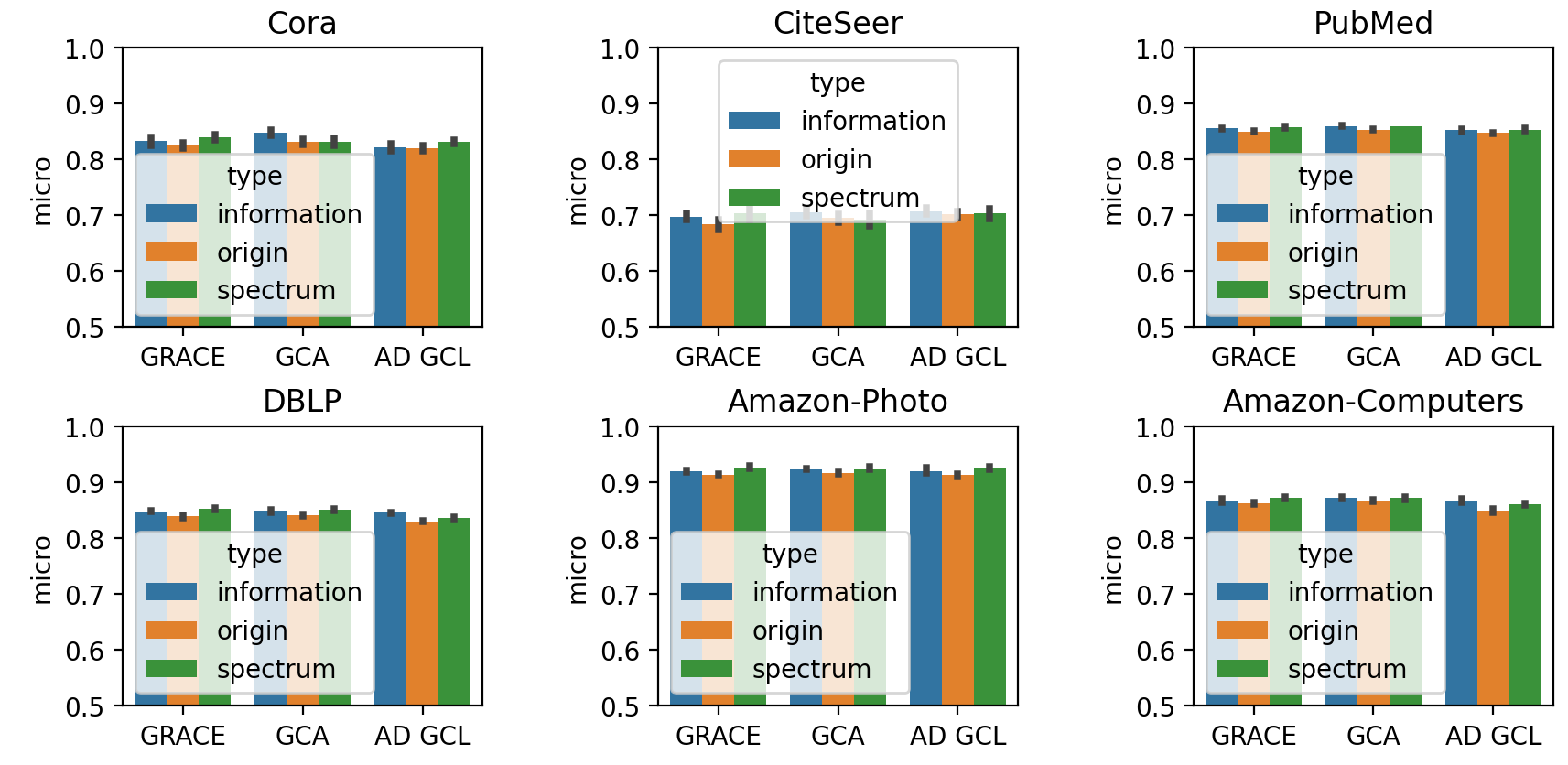}
        \caption{The error bar of algorithms}
        \label{fig:error_bar}
    \end{figure}
\subsection{Hyperparameter Sensitivity}\label{appendix:hyperparameter_sensitivity}
    \begin{figure}
        \centering
        \includegraphics[width=0.9\textwidth]{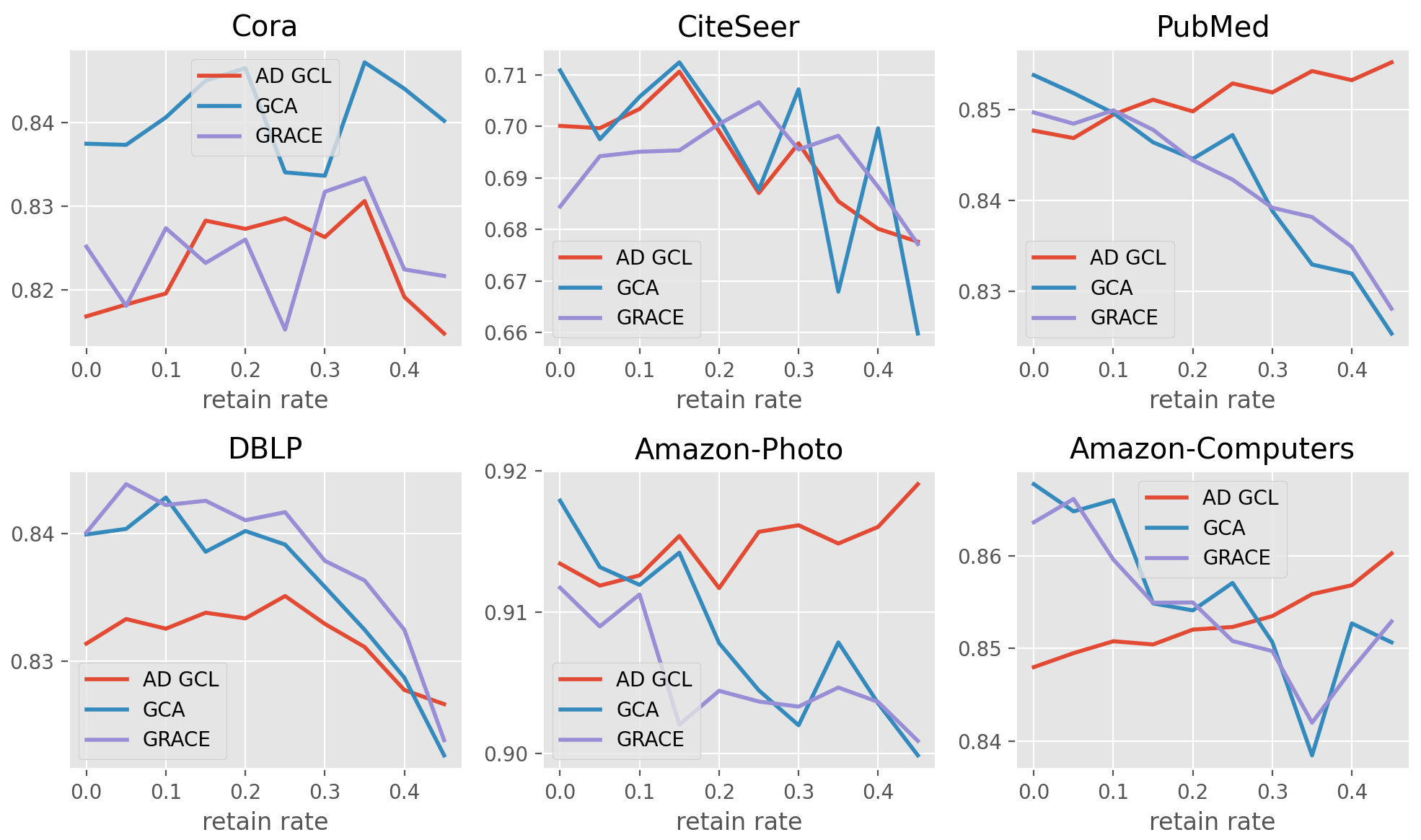}
        \caption{accuracy on downstream tasks with different retain rate}
        \label{fig:hyperparameter_retain_rate}
    \end{figure}

    \begin{figure}
        \centering
        \includegraphics[width=0.9\textwidth]{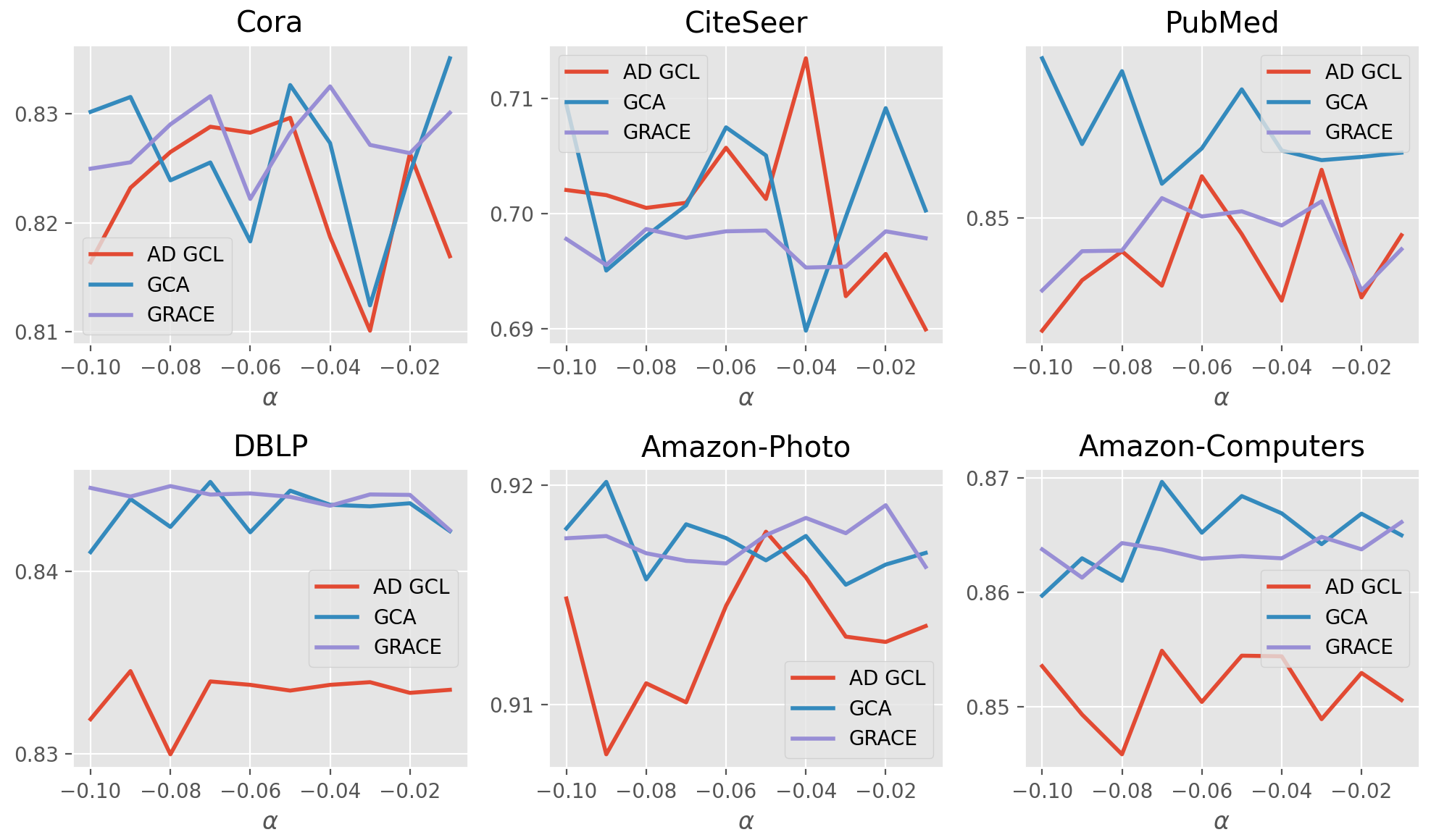}
        \caption{Accuracy on downstream tasks with different $\alpha$}
        \label{fig:hyperparameter_alpha}
    \end{figure}
    \textbf{Analysis of retain rate}. Retain rate controls how many important features/edges we kept, and how many unimportant ones dropped. We can see from Figure \ref{fig:hyperparameter_retain_rate} that AD-GCL benefits from a larger retain rate as it is designed to minimize the mutual information, and lots of vital structures are dropped. And large datasets like PubMed, DBLP benefits less, it is mainly because a graph with more edges are more likely to maintain enough information than graph with little edges. For example, after a 30\% dropout on edges, a graph with 1000 edges would still kept enough information for downstream tasks, but a graph with 10 edges would probably lose some vital information.

    \textbf{Analysis of $\alpha$}. $\alpha$ controls how much $|\lambda|$ will decrease, as we take a step back when the $|\lambda|$ decreases too much, the hyperparameter $\alpha$ does not matter so much. But as shown in Figure \ref{fig:hyperparameter_alpha}, it still performs more steady on large graphs as a wrong modification on a single $\lambda$ matters less than on small graphs.

\subsection{Time Complexity and Error Bar}

    \begin{table}[htbp]
      \centering
      \caption{The time consumption (seconds) of algorithms}
        \begin{tabular}{ccccccc}
        \toprule
        & Cora  & CiteSeer & PubMed & DBLP  & Amazon-P & Amazon-C \\
        \midrule
        GRACE & 8.02  & 10.08 & 62.37 & 56.89 & 19.05 & 28.71 \\
        GRACE+I & 10.74 & 13.49 & 68.97 & 62.8  & 22.67 & 29.61 \\
        GRACE+S & 9.61  & 12.46 & 78.11 & 69.44 & 21.13 & 36.94 \\
        \bottomrule
        \end{tabular}%
      \label{tab:time_efficiency}%
    \end{table}%
    From Table \ref{tab:time_efficiency}, we can observe that the information augmentation method achieve better performance with only few more time consuming, this is mainly because we do not calculate the importance of features/edges every epoch like GCS \citep{GCL_saliency}, we only calculate it once and use the same importance for the following training. However, the spectrum augmentation method consumes more time on large graphs like PubMed and DBLP, this is mainly we explicitly change the spectrum and calculate the new adjacency matrix, which could be replaced by some approximation methods but to prevent interference from random noise and show that Theorem \ref{theorem:spec_NCE} is meaningful, we still conduct eigen decomposition, but it is worth mentioning that the time complexity could be reduced by some approximation methods \citep{GCL_specturm_NCE}.

    The error bar is reported in Figure \ref{fig:error_bar}, the experiments are conducted repeatedly for 10 times, we can observe that both the information augmentation and spectrum augmentation achieve better results, and they performs stably.

\subsection{Combination of Information\&Spectrum Augmentation} \label{appendix:IS_combine}
\begin{table}[htbp]
  \centering
  \caption{Combine the information\&Spectrum Augmentation (GRACE+IS) }
    \begin{tabular}{lllllll}
    \toprule
          & Cora  & CiteSeer & PubMed & DBLP  & Amazon-P & Amazon-C \\
    \midrule
    GRACE & 82.52±0.75 & 70.44±1.49 & 84.97±0.17 & 84.01±0.34 & 91.17±0.15 & 86.36±0.32 \\
    GRACE+I & 83.78±1.08 & 72.89±0.97 & 84.97±0.14 & 84.80±0.17 & 91.64±0.21 & 84.54±0.53 \\
    GRACE+S & 83.61±0.85 & 72.83±0.47 & 85.45±0.25 & 84.83±0.18 & 91.99±0.35 & 87.67±0.33 \\
    GRACE+IS & \textbf{84.58±0.79} & \textbf{72.94±0.52} & \textbf{85.62±0.17} & \textbf{84.87±0.25} & \textbf{92.04±0.32} & \textbf{87.73±0.41} \\
    \bottomrule
    \end{tabular}%
  \label{tab:IS}%
\end{table}%
We combine the information augmentation and spectrum augmentation methods and show the result in Table \ref{tab:IS}. We can observe that combine the two methods achieve the best performance. We can observe that for larger and denser graphs, the information could still be well-preserved even after strong augmentation, rendering the augmentation less powerful compared to smaller graphs. And the spectrum augmentation modify the spectrum based on InfoNCE loss which will be more stable on larger graphs, so it help more significantly in larger graphs.

\section{Related Work} \label{appendix:related_work}

    \textbf{Graph Contrastive Learning.} Graph Contrastive Learning has shown its superiority and lots of researcher are working on it. DGI \citep{DGI} contrasts between local node embeddnig and the global summary vector; GRACE \citep{GRACE}, GCA \citep{GCL_GCA} and GraphCL \citep{GCL_GraphCL} randomly drop edges and features; AD-GCL \citep{GCL_AD_GCL} and InfoGCL \citet{GCL_InfoGCL} learn an adaptive augmentation with the help of different principles. In theoretical perspective, \citet{GCL_specturm_NCE} correlates the InfoNCE loss with graph spectrum, and propose that augmentation should be more focused on high frequency parts. \citet{GCL_VCL_different} further discuss that contrastive learning in graph is different with images. \citet{GCL_max_spectrum} thinks that augmentation maximize the spectrum difference would help, and \citet{GCL_Information} analyse GCL with information theory.

    \textbf{Contrastive Learning Theory.} By linking downstream classification and contrastive learning objectives, \citet{CL_first} propose a theoretical generalization guarantee. \citet{CL_negative_sample_number} further explore how does the number of negative samples influence the generalization. And \citet{CL_good_view,CL_min_sufficient} further discuss what kind of augmentation is better for downstream performance. Then \citet{CL_alignment_divergence} propose that perfect alignment and uniformity is the key to success while \citet{chaos} argues augmentation overlap with alignment helps gathering intra-class nodes by stronger augmentation. However, \citet{CL_inductive_bias} show that augmentation overlap is actually quite rare while the downstream performance is satisfying. So the reason why contrastive learning helps remains a mystery, in this paper we propose that the stronger augmentation mainly helps contrastive learning by separating inter-class nodes, and different from previous works \citep{chaos,CL_alignment_divergence,CL_hwr}, we do not treat perfect alignment as key to success, instead a stronger augmentation that draw imperfect alignment could help.

\end{document}

%% file: math_commands.tex
\usepackage{amsmath,amsfonts,bm}

\def\eqref#1{equation~\ref{#1}}

\def\1{\bm{1}}

\def\vx{{\bm{x}}}

\def\mA{{\bm{A}}}

\def\mD{{\bm{D}}}

\def\mF{{\bm{F}}}

\def\mM{{\bm{M}}}

\def\mS{{\bm{S}}}

\def\mX{{\bm{X}}}

\DeclareMathAlphabet{\mathsfit}{\encodingdefault}{\sfdefault}{m}{sl}
\SetMathAlphabet{\mathsfit}{bold}{\encodingdefault}{\sfdefault}{bx}{n}

\def\gE{{\mathcal{E}}}

\def\gG{{\mathcal{G}}}

\def\gL{{\mathcal{L}}}

\def\gV{{\mathcal{V}}}

\def\sG{{\mathbb{G}}}

\newcommand{\E}{\mathbb{E}}

\newcommand{\R}{\mathbb{R}}

